\documentclass{article}

\usepackage{microtype}
\usepackage{graphicx}
\usepackage{subfigure}
\usepackage{booktabs} % for professional tables

\usepackage{hyperref}

\usepackage[accepted]{icml2024}

\usepackage{amsmath}
\usepackage{amssymb}
\usepackage{mathtools}
\usepackage{amsthm}

\usepackage[capitalize,noabbrev]{cleveref}

\theoremstyle{plain}

\theoremstyle{definition}

\theoremstyle{remark}

\usepackage[textsize=tiny]{todonotes}

\usepackage{amsfonts}
\usepackage{bm}
\usepackage{xspace}
\usepackage{dsfont}
\usepackage{hyperref}
\usepackage{graphicx}
\usepackage{color}
\usepackage{amsthm}
\usepackage{subfigure}
\usepackage{mathrsfs}
\usepackage{enumitem}
\usepackage{bbm}
\usepackage{wrapfig}
\usepackage{makecell}
\usepackage{colortbl}
\usepackage{booktabs}
\usepackage{pifont}

\usepackage{algorithm, algorithmic}
\newcommand\RCOMMENT[1]{\hfill\(\triangleright\) #1}
\usepackage{ifthen}
\newcounter{subroutine}


\setboolean{ALC@noend}{true}

\usepackage{multirow}
\usepackage{float}
\usepackage[nottoc]{tocbibind}
\usepackage{dcolumn}
\usepackage{longtable}
\newcommand{\eps}{\varepsilon}
\newcommand{\Jstd}{J_\textrm{std}(\theta_c)}
\newcommand{\Jrisk}{J_\textrm{risk}(\theta_c; \eps)}
\newcommand{\Jentropy}{J_\textrm{entropy}(\theta_c; \eps)}
\newcommand{\dysymnet}{\textsc{DySymNet}\xspace}
\newcommand{\ntest}{N_\text{test}}
\newcommand{\placeholder}{\mathord{\color{black!33}\bullet}}
\definecolor{C0}{HTML}{1f77b4}
\definecolor{C1}{HTML}{ff7f0e}
\definecolor{C2}{HTML}{2ca02c}
\definecolor{C3}{HTML}{d62728}
\definecolor{C4}{HTML}{9467bd}
\definecolor{C5}{HTML}{8c564b}
\definecolor{C6}{HTML}{e377c2}
\definecolor{C7}{HTML}{7f7f7f}
\definecolor{C8}{HTML}{bcbd22}
\definecolor{C9}{HTML}{17becf}
\definecolor{C10}{HTML}{1F77B4}
\definecolor{C11}{HTML}{FF7F0F}
\definecolor{C12}{HTML}{E5F2E5}
\definecolor{C13}{HTML}{EB9394}

\icmltitlerunning{A Neural-Guided Dynamic Symbolic Network for Exploring Mathematical Expressions from Data}

\begin{document}

\twocolumn[
\icmltitle{A Neural-Guided Dynamic Symbolic Network for Exploring Mathematical Expressions from Data}



\icmlsetsymbol{cor}{*}

\begin{icmlauthorlist}
\icmlauthor{Wenqiang Li}{aaa,bbb,ccc}
\icmlauthor{Weijun Li}{aaa,bbb,ccc,ddd,cor}
\icmlauthor{Lina Yu}{aaa,bbb,ccc}
\icmlauthor{Min Wu}{aaa,bbb,ccc}
\icmlauthor{Linjun Sun}{aaa,bbb,ccc}
\icmlauthor{Jingyi Liu}{aaa,bbb,ccc}
\icmlauthor{Yanjie Li}{aaa,bbb,ccc}
\icmlauthor{Shu Wei}{aaa,bbb,ccc}
\icmlauthor{Yusong Deng}{aaa,bbb,ccc}
\icmlauthor{Meilan Hao}{aaa,bbb,ccc}

\end{icmlauthorlist}

\icmlaffiliation{aaa}{AnnLab, Institute of Semiconductors, Chinese Academy of Sciences, Beijing, China}
\icmlaffiliation{bbb}{School of Electronic, Electrical and Communication Engineering \& School of Integrated Circuits, University of Chinese Academy of Sciences, Beijing, China}
\icmlaffiliation{ccc}{Beijing Key Laboratory of Semiconductor Neural Network Intelligent Sensing and Computing Technology, Beijing, China}
\icmlaffiliation{ddd}{Center of Materials Science and Optoelectronics Engineering, University of Chinese Academy of Sciences, Beijing, China}

\icmlcorrespondingauthor{Weijun Li}{wjli@semi.ac.cn}

\icmlkeywords{Machine Learning, symbolic regression, AI for science}

\vskip 0.3in
]


\printAffiliationsAndNotice{*Corresponding author.}  % leave blank if no need to mention equal contribution

\begin{abstract}
Symbolic regression (SR) is a powerful technique for discovering the underlying mathematical expressions from observed data. Inspired by the success of deep learning, recent deep generative SR methods have shown promising results. However, these methods face difficulties in processing high-dimensional problems and learning constants due to the large search space, and they don't scale well to unseen problems.
In this work, we propose \dysymnet, a novel neural-guided \textbf{Dy}namic \textbf{Sym}bolic \textbf{Net}work for SR. Instead of searching for expressions within a large search space, we explore symbolic networks with various structures, guided by reinforcement learning, and optimize them to identify expressions that better-fitting the data. 
Based on extensive numerical experiments on low-dimensional public standard benchmarks and the well-known SRBench with more variables, \dysymnet shows clear superiority over several representative baseline models. Open source code is available at \url{https://github.com/AILWQ/DySymNet}.

\end{abstract}

\begin{figure*}[htbp]
    \centering
    \includegraphics[width=\textwidth]{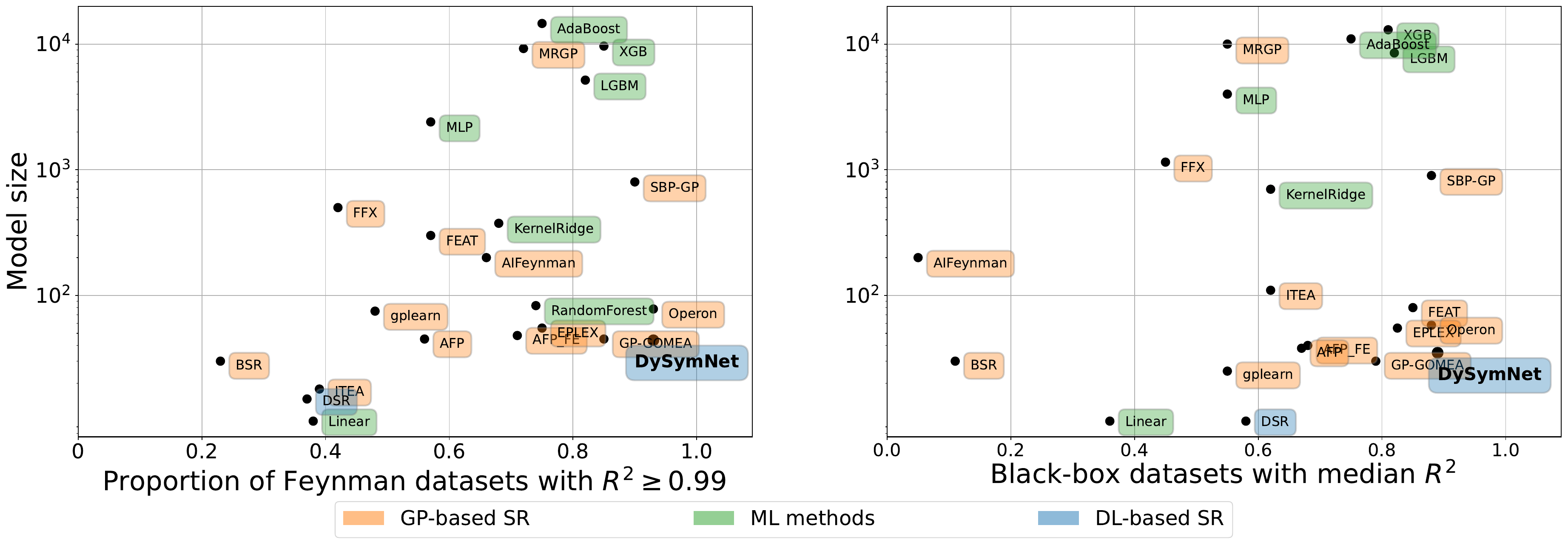}
    \caption{\textbf{\dysymnet outperforms previous DL-based and GP-based SR methods in terms of fitting accuracy while maintaining a relatively small symbolic model size.} Pareto plot comparing the average test performance and model size of our method with baselines provided by the SRBench benchmark~\citep{la1contemporary}, both on \textit{Feynman} dataset (left) and \textit{Black-box} dataset (right). We use the colors to distinguish three families of models: \textbf{\textcolor{C0}{deep-learning based SR}},  \textbf{\textcolor{C1}{genetic programming-based SR}} and  \textbf{\textcolor{C2}{classic machine learning methods}} (which do not provide interpretable solutions).}
    \label{fig:perato}
\end{figure*}

\section{Introduction}
Numerous phenomena in the natural world, such as physical laws, can be precisely described using mathematical expressions. 
Symbolic regression (SR) is an effective machine-learning technique that involves discovering mathematical expressions that describe a dataset with accuracy. Unlike polynomial or neural network-based regression, SR aims to unveil the fundamental principles underlying the data generation process. This method is analogous to how physicists use explicit mathematical models to explain physical phenomena. For instance, Isaac Newton's laws of motion provided a mathematical framework for describing object motion, while Albert Einstein's theory of relativity introduced new equations to explain the behavior of objects in motion. More specifically, given a dataset $(X, y)$, where each feature $X_i \in \mathbb{R}^n$ and target $y_i \in \mathbb{R}$, the goal of SR is to identify a function $f$ (i.e., $y \approx f(X)$ $: \mathbb{R}^n \rightarrow \mathbb{R}$) that best ﬁts the dataset, where the functional form of $f$ is a short closed-form mathematical expression.

SR is a challenging task because the search space of expressions grows exponentially with the length of the expression, while the position and value of numeric further exacerbate its difficulty. The traditional SR methods mainly involve heuristic search methods based on genetic programming (GP)~\citep{forrest1993genetic, koza1994genetic, schmidt2009distilling, staelens2013constructing, arnaldo2015building, blkadek2019solving}. They represent the expression as a binary tree and find high-fitness solutions through iterative evolutions in the large functional search space. While GP-based SR methods have the capability to solve nonlinear problems, they often yield complex expressions and are computationally expensive. Moreover, these methods are known to exhibit high sensitivity to hyperparameters, which can complicate the optimization process. 

A more recent line of research has made use of the neural network to tackle the aforementioned shortcomings. \citet{sahoo2018learning} proposed an equation learner (EQL), a fully-connected network where elementary functions are used as activation functions. They try to constrain the search space by optimizing a pre-defined network. The limitation of EQL is that the pre-defined architecture of the network limits the complexity of the predicted expression and is not flexible for different specific problems. Recently, reinforcement learning (RL)-based methods~\citep{petersen2020deep, mundhenk2021symbolic} for SR have shown promising results. They directly search expressions in the large functional space guided by RL. Although it is effective in dealing with low-dimensional problems without constants, they face difficulties in handling high-dimensional problems and constant optimization due to the large search space. Inspired by the success of large-scale pre-training, there has been a growing interest in the SR community for transformer-based models~\citep{valipour2021symbolicgpt, biggio2021neural, kamienny2022end, litransformer}. These approaches are inductive: they are pre-trained on a large-scale dataset to generate a pre-order traversal of the expression tree in a single forward pass for any new dataset. Thus, transformer-based SR methods possess the advantage of generating expressions quickly. However, they may encounter problems in inference when the given data is out-of-distribution compared to the synthetic data, and can not generalize to unseen input variables of a higher dimension from those seen during pre-training.

Overall, the current mainstream SR methods are mainly generating expression trees from scratch or experience. 
EQL provides a good idea for reducing the search space and inferencing high-dimensional problems, despite the lack of flexibility.
To address these issues, we propose \dysymnet, a novel neural-guided \textbf{Dy}namic \textbf{Sym}bolic \textbf{Net}work for SR. 
We use a controller for sampling various architectures of symbolic network guided by policy gradients, which reduce the vast search space of expressions to that of symbolic network structures while retaining powerful formulaic representation capability. 

In summary, we introduce the main contributions in this work as follows:
\begin{itemize}[leftmargin=*]
    \item Proposed \dysymnet offers a new search paradigm for SR that searches the symbolic network with various architectures instead of searching expression trees in the vast functional space.

    \item Proposed \dysymnet is adaptable and can converge to a compact and interpretable symbolic expression, with promising capabilities for effectively solving high-dimensional problems that are lacking in mainstream SR methods.

    \item We demonstrated that \dysymnet outperforms several representative baselines across various SR standard benchmarks and the well-known SRBench with more variables. Its state-of-the-art performance surpasses that of the baselines in terms of accuracy and robustness to noise.

\end{itemize}

% \begin{figure*}[h!]
%     \centering
%     \includegraphics[width=\textwidth]{figures/overview-new.pdf}
%     \caption{\textbf{Overview of neural-guided \dysymnet.} First, we sample batch descriptions of \dysymnet architecture autoregressively via RNN. Then, we instantiate and train \dysymnet through backpropagation and weight pruning. Finally, we use BFGS to refine the constants and train RNN via risk-seeking policy gradient with entropy.}
%     \label{fig:overview}
% \end{figure*}

\section{Related Work}
\paragraph{Symbolic regression from scratch}
SR methods can be broadly classified into two categories: the first category comprises methods that start from scratch for each instance, while the second category involves transformer methods based on large-scale supervised learning. Traditionally, genetic programming (GP) algorithms~\citep{forrest1993genetic} are commonly employed to search the optimal expression for given observations~\citep{koza1994genetic, dubvcakova2011eureqa}. However, these methods tend to increase in complexity without much performance improvement, and it is also problematic to tune expression constants only by using genetic operators. Recently, the neural networks were used for SR. ~\citet{martius2016extrapolation} leverage a pre-defined fully-connected neural network to identify the expression, which constrains the search space but is not flexible. ~\citet{petersen2020deep} propose deep symbolic regression (DSR), using reinforcement learning (RL) to guide a policy network to directly output a pre-order traversal of the expression tree. Based on DSR, a combination of GP and RL~\citep{mundhenk2021symbolic} was presented, where the policy network is used to seed the GP's starting population. ~\citet{xu2023rsrm} proposed a Reinforcement Symbolic Regression Machine (RSRM) that masters the capability of uncovering complex math expressions from only scarce data. They employ a Monte Carlo tree search (MCTS) agent and combined with RL to explore optimal expression trees based on measurement data.
Although the promising results make it the currently recognized state-of-the-art approach to SR tasks. Nevertheless, the limitations of this method and others from scratch are obvious, namely, the large search space of the expression trees makes them face difficulties in handling high-dimensional problems and constants optimization.
\paragraph{Transformer-based model for symbolic regression}
In recent years, the transformer~\citep{vaswani2017attention} has gained considerable attention in the field of natural language processing. For instance, in machine translation, the transformer model has been extensively employed due to its remarkable performance. Recently, transformer-based models have been highly anticipated for SR. For example, Valipour \textit{et al.}~\citep{valipour2021symbolicgpt} proposed SymbolicGPT that models SR as a machine translation problem. They established a mapping between the data point space and the expression space by encoding the data points and decoding them to generate an expression skeleton on the character level. Similarly, Biggo \textit{et al.}~\citep{biggio2021neural} introduces NeSymReS that can scale with the amount of synthetic training data and generate expression skeletons. Kamienny \textit{et al.}~\citep{kamienny2022end} proposed an end-to-end framework based on the transformer that predicts the expression along with its constants. These methods encode the expression skeleton or the complete expression into a sequence, corresponding to the pre-order traversal of the expression tree, then trained on token-level cross-entropy loss. Despite showing promising results, their training approach suffers from the problem of insufficient supervised information for SR. This is because the same expression skeleton can correspond to multiple expressions with different coefficients, leading to ill-posed problems. To address this issue, ~\citet{litransformer} proposed a joint supervised learning approach to alleviate the ill-posed problem. Overall, transformer-based model for SR has a clear advantage in inference speed compared to methods starting from scratch. 
Additionally, researchers have made efforts to improve the accuracy of Transformer-based SR methods in searching for expressions~\citep{holt2022deep,liu2023snr,wu2023discovering,li2024discovering}.
However, their limitations are poor flexibility and model performance dependence on the training set. They may encounter problems when inferring expressions outside the training set distribution.

\begin{figure*}[h!]
    \centering
    \includegraphics[width=\textwidth]{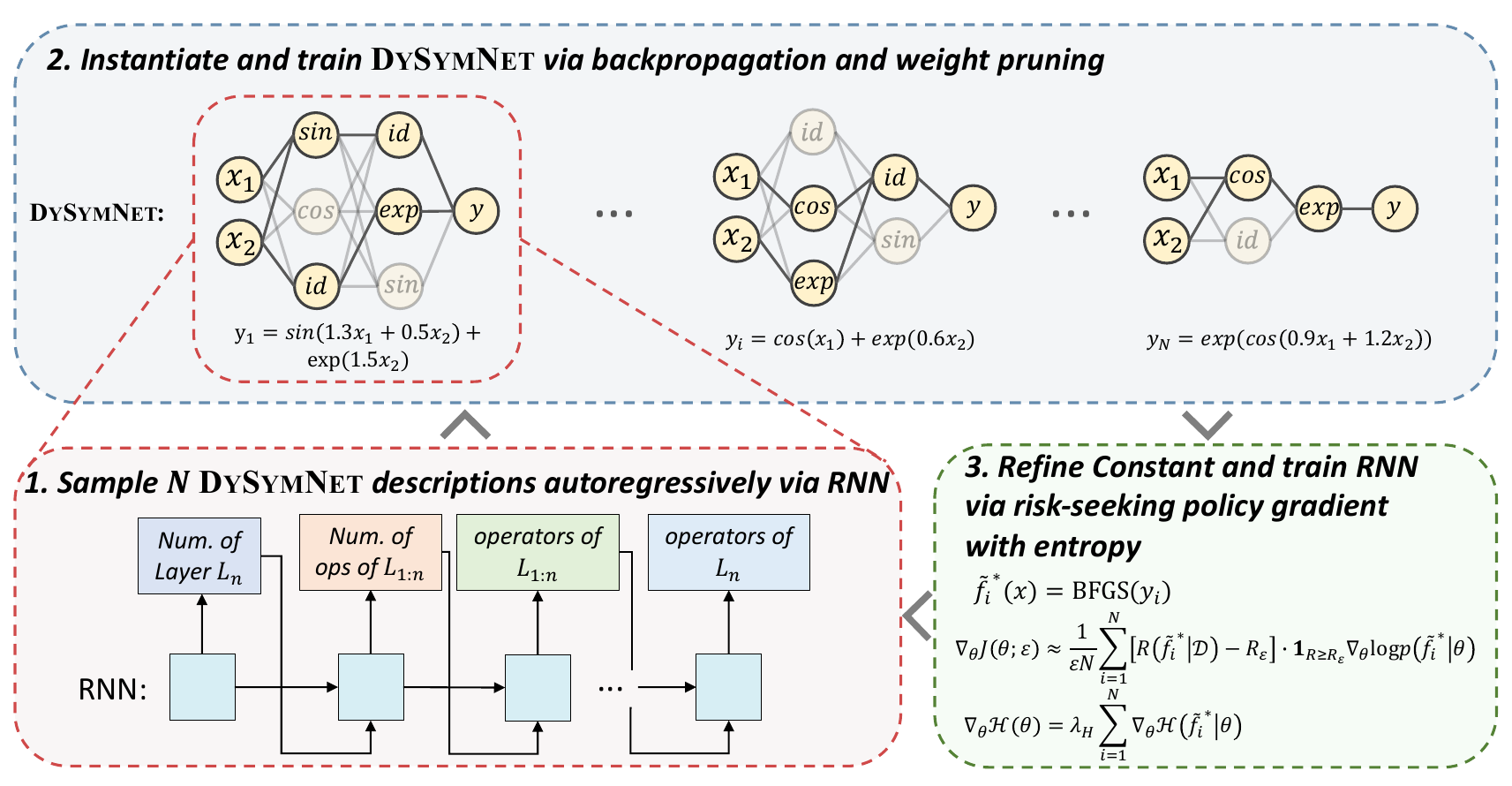}
    \caption{\textbf{Overview of neural-guided \dysymnet.} First, we sample batch descriptions of \dysymnet architecture autoregressively via RNN. Then, we instantiate and train \dysymnet through backpropagation and weight pruning. Finally, we use BFGS to refine the constants and train RNN via risk-seeking policy gradient with entropy.}
    \label{fig:overview}
\end{figure*}
\section{Methodology}
The overall approach is visualized in Figure~\ref{fig:overview}. Pseudocode for \dysymnet is shown in Appendix~\ref{app:code}.

\subsection{Identify expression from \dysymnet}
\label{sub:identify}

\paragraph{\dysymnet architecture}
In this work, \dysymnet is flexible and adaptable in architecture, rather than being fixed and unchanging. The architecture of the symbolic network is controlled by a recurrent neural network (RNN), with further details explained in Section~\ref{sec:gen}. Each symbolic network is a fully connected feed-forward network with units representing the building blocks of algebraic expressions. Each layer of the \dysymnet is automatically designed by the RNN for a specific SR task, and it is subject to \textit{evolve} over time. In a \dysymnet with $L$ layers, there are $L-1$ hidden layers that consist of a linear mapping followed by non-linear transformations. The linear mapping at level $l$ maps the $n'$-dimensional input $\bm{h}^{(\ell-1)}$ to the $m$-dimensional intermediate representation $\bm{z}$ given by
\begin{equation}
    \bm{z}^{(\ell)} = \bm{W}^{(\ell)}\bm{h}^{(\ell-1)} + \bm{b}^{(\ell)}, 
    \nonumber
\end{equation}
where $\bm{h}^{(\ell-1)}\in\mathbb{R}^{n'}$ is the output of the previous layer, with the convention $\bm{h}^{(0)}=\bm{x}$. The weight matrix $\bm{W}^{(\ell)} \in \mathbb{R}^{m\times n'}$ and the bias vector $\bm{b}^{(\ell)} \in \mathbb{R}^d$ are free parameters that learned during training. In practice, we predefine a function library $\mathcal{F}$ of available operators, e.g., $\{\textrm{Id}, +, -, \times, \sin(\cdot), \cos(\cdot), \exp(\cdot) \}$, to be selected in generating \dysymnet. Suppose that we have presently instantiated a specific symbolic network, whereby its $\ell$-th layer contains $u$ unary operator units, $f_i: \mathbb{R} \to \mathbb{R}$, for $i=1,\dots,u$, and $v$ binary operator units, $g_i: \mathbb{R} \times \mathbb{R} \to \mathbb{R}$ for $j=1,\dots,v$. The $\ell$-th layer output $\bm{h}^{(\ell)}$ is formed by the concatenation of units outputs
\begin{equation}
\begin{aligned}
\bm{h}^{(\ell)}:= & \left(f_1\left(\bm{z}_1^{(\ell)}\right), f_2\left(\bm{z}_2^{(\ell)}\right), \ldots, f_u\left(\bm{z}_u^{(\ell)}\right)\right., \\
& \left.g_1\left(\bm{z}_{u+1}^{(\ell)}, \bm{z}_{u+2}^{(\ell)}\right), \ldots, g_v\left(\bm{z}_{u+2 v-1}^{(\ell)}, \bm{z}_{u+2v}^{(\ell)}\right)\right).
\end{aligned}
\nonumber
\end{equation}
Specifically, the non-linear transformation stage has $n=u+v$ outputs and $m=u+2v$ inputs. The unary units, $f_1, \dots, f_u$, $f_i \in \{\textrm{Id}, \sin(\cdot), \cos(\cdot), \exp(\cdot)\}$, receive the respective component, $z_1, \dots, z_u$ as inputs. The binary units $g_1, \dots, g_v$, $g_j \in \{+, -, \times, \div \}$, receive the remaining component, $z_{u+1}, \dots, z_{u+2v}$, as input in pairs of two. For example, the multiplication unit computes the product of their two input values: $g_j(\bm{z}_{u+2j-1}, \bm{z_}{u+2j}) := \bm{z}_{u+2j-1} \cdot \bm{z}_{u+2j}$, for $j=1,\dots, v.$

The last layer computes the regression values by a linear read-out without any operator units
\begin{equation}
    \bm{y}^{(L)} := \bm{W}^{(L)} \bm{h}^{(L)} + \bm{b}^{(L)}.
    \nonumber
\end{equation}
The architecture of the symbolic network is illustrated in Figure~\ref{fig:overview}. We define the expression denoted from the particular symbolic network  as $\tilde{f}(\cdot)$.

\paragraph{\dysymnet training}
Similar to conventional fully-connected neural networks, the symbolic network possesses its own set of free parameters, denoted as $\bm{\Theta}=\{\bm{W}^{(1)}, \dots, \bm{W}^{(L)}, \bm{b}^{(1)}, \dots, \bm{b}^{(L)}\}$, which is predetermined and can undergo end-to-end training through back-propagation.
While it may be possible to fit a more accurate expression that incorporates a greater number of terms, for a successful SR method, the goal is to find the simplest expression that accurately explains the dataset. This requires a trade-off between accuracy and complexity. To achieve this, two approaches are adopted. First, similar to the transformer-based methods that impose constraints on the pre-order traversal length of the expression tree, we limit the maximum depth of the symbolic network to 5 layers and the maximum width to 6 neurons prior to training. Second, regularization techniques are applied during training to sparsify the weights, followed by pruning according to specific rules to obtain a simplified mathematical expression that accurately explains the dataset. More specifically, given the dataset $\mathcal{D}=\{(\bm{x_i}, y_i)\}_{i=1}^n$, the training objective is then given in the following:
\begin{equation}
    \mathcal{L} = \frac{1}{n} \sum^{n}_{i=1}\|\tilde{f}(\bm{x}_i)-y_i\|^2 + \lambda \sum^{L}_{\ell=1} |\bm{\Theta}^{(\ell)}|_r
    \nonumber
\end{equation}
where $n$ indicates the size of training data, the last term represents the regularization term, $\lambda$ is a switch factor of the regularization term, with a value of $\{0, 1\}$, and $L$ is the number of hidden layers.

Furthermore, we propose an adaptive gradient clipping strategy to enhance the stability of the training process. Specifically, the gradient norm at each time step is calculated as the moving average of the gradient norm within a certain time window. Then, this average value is used to adjust the threshold $\gamma$ for gradient clipping, which is applied to the gradients of \dysymnet. The gradient clipping threshold with window size $w$ and weight factor $c$ during the $n$-th epoch of training \dysymnet is given by: 
$\gamma = \frac{c}{w}\sum_{i=1}^w \sum_{\ell=1}^L \Vert \Theta^{(\ell)} \Vert_2$,
where we set the window size $w$ is 50 and the weight factor $c$ is 0.1. In the ablation experiments in Section~\ref{sec:abla}, we demonstrated the effectiveness of this technique.

\paragraph{Regularization and prune}
The purpose of regularization is to obtain a sparse symbolic network, which is consistent with the goal of SR to obtain an interpretable and concise expression. The ideal scenario would involve directly minimizing the $L_0$ norm, wherein the regularization term penalizes the presence of non-zero weights irrespective of their magnitude, effectively pushing the solution towards sparse representations. However, the utilization of $L_0$ regularization presents a combinatorial conundrum that is NP-hard to solve, thus rendering it incompatible with gradient descent methods that are typically employed for optimizing neural networks. 
$L_1$ regularization has long been recognized as a prominent sparsity technique, which is used in the original EQL~\citep{sahoo2018learning}. Despite its effectiveness in promoting sparsity, since $L_1$ regularization also penalizes the magnitude of the weights, $L_{0.5}$ has been proposed to enforce sparsity more strongly without penalizing the magnitude of the weights as much as $L_1$~\citep{xu20101, zong2012representative}. Following~\citep{kim2020integration}, we use a smoothed version of $L_{0.5}^*$ proposed in~\citep{wu2014batch} that has been shown to avoid singularity in the gradient, a challenge often encountered in gradient descent-based optimization when the weights approach zero. The $L_{0.5}^*$ regularizer employs a piecewise function to smooth out the function at small magnitudes:
\begin{equation}
L_{0.5}^*(w)= \begin{cases}|w|^{1 / 2} & |w| \geq a \\ \left(-\frac{w^4}{8 a^3}+\frac{3 w^2}{4 a}+\frac{3 a}{8}\right)^{1 / 2} & |w| < a\end{cases}, 
\nonumber
\end{equation}
where $a \in \mathbb{R}^+$ is the transition point between the standard $L_{0.5}$ function and the smoothed function. After training, we keep all weights $w \in \bm{W}^{1\dots L}$ that are close to 0 at 0, i.e., if $|w| < 0.01$ then $w=0$ to obtain the sparse structure.

\subsection{Generate \dysymnet  with a controller recurrent neural network}
\label{sec:gen}
We leverage a controller to generate architecture descriptions of \dysymnet. The controller is implemented as an RNN. We can use the RNN to generate architecture parameters as a sequence of tokens sequentially. Once the RNN finishes generating an architecture, a \dysymnet with this architecture is instantiated and trained. At convergence, we obtain the compact symbolic expression and evaluate it with a reward function. The reward function is not differentiable with respect to the parameters of the controller RNN, $\theta_c$; thus, we optimize $\theta_c$ via policy gradient in order to maximize the expected reward of the proposed architectures. Next, we will elaborate on the specific process of generating architecture parameters of \dysymnet using the RNN, the reward function definition, and the training algorithm.

\paragraph{Generative process}
We adopt a Markov Decision Process (MDP)~\citep{sutton2018reinforcement} modeling the process of generating architectures. An RL setting consists of four components ($\mathcal{S}, \mathcal{A}, \mathcal{P}, \mathcal{R}$) in a MDP. 
In this view, the list of tokens that the controller RNN predicts can be viewed as a list of actions $a_{1:T}\in \mathcal{A}$ to design an architecture for a \dysymnet. Hence, the process of generating architectures can be framed in RL as follows: the agent (RNN) emits a distribution over the architecture $p(a_{1:T}|\theta_c)$, observes the environment (current architecture) and, based on the observation, takes an action (next available architecture parameter) and transitions into a new state (new architecture). The architecture parameters of the \dysymnet mainly consist of three parts: \textbf{the number of layers}, \textbf{the number of operators for each layer}, and \textbf{the type of each operator}. As shown in Figure~\ref{fig:overview}, for a specific process, the RNN first samples the number of layers in the network. Then, for each layer, the RNN samples the number of operators in the $i$-th layer and the type of operators sequentially. At each time step, the input vector $\bm{x}_t$ of the RNN is obtained by embedding the previous parameter $a_{t-1}$. Each episode refers to the complete process of using the RNN to sample a \dysymnet, instantiating it, and training it to obtain a symbolic expression. This process is performed iteratively until the resulting expression meets the desired performance criteria.

\paragraph{Reward definition}
As the expression $\tilde{f}(\cdot)$ identified from the \dysymnet can only be evaluated at the end of the episode, the reward $R$ equals 0 until the final step is reached. Then, we use the nonlinear optimization algorithm BFGS~\citep{1984Practical} to refine the constants to get a better-fitting expression $\tilde{f}^*(\cdot)$. The constants in the expression $\tilde{f}(\cdot)$ are used as the initial values for the BFGS algorithm. To align with the training objective of the \dysymnet, we first calculate the standard fitness measure mean squared error (MSE) between the ground-truth target variable $y$ and the predicted target variable $\tilde{y}=\tilde{f}^*(\bm{x})$. That is, given a dataset $\mathcal{D}=\{(\bm{x_i}, y_i)\}_{i=1}^n$ of size $n$ and candidate expression $\tilde{f}^*(\cdot)$, $\textrm{MSE}(\tilde{f}^*(\bm{x}), y) = \frac{1}{n}\sum^n_{i=1}(\tilde{f}^*(\bm{x}_i) - y_i)^2$.
Then, to bound the reward function, we apply a squashing function: $R(\tilde{f}^*|\mathcal{D}) = {1} / (1 + \textrm{MSE})$.

\paragraph{Training the RNN using policy gradients}
We use the accuracy of $\tilde{f}^*(\cdot)$ as the reward signal and use RL to train the controller RNN. More concretely, to find the optimal architecture of \dysymnet, we first consider the standard REINFORCE policy gradient~\citep{williams1992simple} objective to maximize $\Jstd$, defined as the expectation of a reward function $R(\tilde{f}^*|\mathcal{D})$ under expressions from the policy: 
\begin{align*}
    J_{\textrm{std}}(\theta_c) \doteq \mathbb{E}_{\tilde{f} \sim p(a_{1:T} | \theta_c)}\left[R(\tilde{f}^*|\mathcal{D})\right]
\end{align*}

The standard policy gradient objective, $\Jstd$, is the desired objective for control problems in which one seeks to optimize the average performance of a policy. However, the final performance in domains like SR is measured by the single or few best-performing samples found during training. For SR, $\Jstd$ is not an appropriate objective, as there is a mismatch between the objective being optimized and the final performance metric. To address this disconnect, we adopt the \textit{risk-seeking policy gradient} proposed in~\citep{petersen2020deep}, with a new learning objective that focuses on learning only on \textit{maximizing best-case performance}. The learning objective $\Jrisk$ is parameterized by $\eps$:
\begin{align}
\label{eqn:J}
\Jrisk \doteq \mathbb{E}_{\tilde{f} \sim p(a_{1:T} | \theta_c)}\left[ R(\tilde{f}^*|\mathcal{D}) \mid R(\tilde{f}^*|\mathcal{D}) \geq R_{\eps}(\theta_c) \right],
\nonumber
\end{align}
where $r_{\eps}(\theta_c)$ is defined as the $(1-\eps)$-quantile of the distribution of rewards under the current policy. This objective aims to increase the reward of the top $\eps$ fraction of samples from the distribution, without regard for samples below that threshold. 
Lastly, in accordance with the maximum entropy RL framework~\citep{haarnoja2018soft}, we add an entropy term $\mathcal{H}$ weighted by $\lambda_{\mathcal{H}}$, to encourage exploration:
\begin{equation}
    \Jentropy \doteq \mathbb{E}_{\tilde{f} \sim p(a_{1:T} | \theta_c)}[\mathcal{H}(\tilde{f}|\theta_c) \mid R(\tilde{f}^*|\mathcal{D}) \geq R_{\eps}(\theta_c)].
    \nonumber
\end{equation}

\section{Experimental Settings}
In this section, we present our experimental settings and results. We evaluate our proposed \dysymnet by answering the following research questions (\textbf{RQs}):

\textbf{RQ1:} Does \dysymnet perform better than other SR algorithms on high-dimensional problems? 

\textbf{RQ2:} Do individual components of \dysymnet contribute to overall performance?

\textbf{RQ3:} Does \dysymnet provide better robustness to noise than other SR algorithms?

\textbf{RQ4:} Can \dysymnet be used in real life to discover the physical laws?

\subsection{Metrics}
We assess our method and baselines using the coefficient of determination ($R^2$): 
\begin{equation}
        R^{2}=1-\frac{\sum_{i}^{\ntest}\left(y_{i}-\tilde{y}_{i}\right)^{2}}{\sum_{i}^{\ntest}\left(y_{i}-\bar{y}\right)^{2}},
        \label{eq:metrics}
        \nonumber
\end{equation}
where $\ntest$ is the number of observations, $y_i$ is the ground-truth value for the $i$-th observation, $\tilde{y}_i$ is the predicted value for the $i$-th observation, and $\bar{y}$ is the averaged value of the ground-truth. $R^2$ measures the goodness of fit of a model to the data, and it ranges from 0 to 1, with higher values indicating a better fit between the model and the data. 

\begin{table*}[h!]
\caption{Average $R^2$ of \dysymnet compared with current strong baselines on the standard benchmarks $(d\leq2)$ and SRBench ($d\geq2$)~\citep{la1contemporary} across 20 independent training runs, where $d$ represents the dimensionality of the problem. 95\% conﬁdence intervals are obtained from the standard error between mean $R^2$ on each problem set.}
\label{tab:fit acc}
\vskip 0.1in
\centering
\resizebox{\textwidth}{!}{
\begin{tabular}{clccccc}
\toprule
\multirow{2}{*}{Data Group}                   & \multirow{2}{*}{Benchmark} & \multicolumn{1}{c}{\dysymnet} & \multicolumn{1}{c}{uDSR} & \multicolumn{1}{c}{NGGP} & \multicolumn{1}{c}{NeSymReS} & \multicolumn{1}{c}{EQL}\\ 
\cmidrule(lr){3-3} \cmidrule(lr){4-4} \cmidrule(lr){5-5} \cmidrule(lr){6-6} \cmidrule(lr){7-7}
                                              &                            & $R^2\uparrow$     & $R^2\uparrow$       & $R^2\uparrow$         & $R^2\uparrow$        & $R^2\uparrow$        \\ 
\cmidrule(lr){1-1} \cmidrule(lr){2-2} \cmidrule(lr){3-3} \cmidrule(lr){4-4} \cmidrule(lr){5-5} \cmidrule(lr){6-6} \cmidrule(lr){7-7}
\multirow{7}{2cm}{\centering Standard $(d \leq 2)$}
                                              & Nguyen                     & $\textbf{0.9999}_{\pm 0.0001}$        & $0.9989_{\pm 0.0000}$         & $0.9848_{\pm 0.0000}$              & $0.9221_{\pm 0.0000}$        & $0.9758_{\pm 0.0032}$     \\
\multicolumn{1}{c}{}                          & Nguyen*                    & $\textbf{0.9999}_{\pm 0.0000}$        & $\textbf{0.9999}_{\pm 0.0000}$  & $\textbf{0.9999}_{\pm 0.0001}$    & $0.3523_{\pm 0.0000}$     & $0.7796_{\pm 0.0019}$  \\
\multicolumn{1}{c}{}                          & Constant                   & $\textbf{0.9992}_{\pm 0.0000}$       & $0.9989_{\pm 0.0002}$         & $0.9989_{\pm 0.0002}$            & $0.7670_{\pm 0.0000}$        & $0.9817_{\pm 0.0088}$ \\
\multicolumn{1}{c}{}                          & Keijzer                    & $\textbf{0.9986}_{\pm 0.0002}$         & $0.9968_{\pm 0.0000}$         & $0.9912_{\pm 0.0005}$           & $0.9358_{\pm 0.0000}$      & $0.9446_{\pm 0.0014}$ \\
\multicolumn{1}{c}{}                          & Livermore                   & $\textbf{0.9896}_{\pm 0.0001}$       & $0.9670_{\pm 0.0003}$        & $0.9665_{\pm 0.0005}$         & $0.9195_{\pm 0.0000}$         & $0.7496_{\pm 0.0002}$  \\
\multicolumn{1}{c}{}                          & R                          & $0.9895_{\pm 0.0012}$               & $\textbf{0.9999}_{\pm 0.0000}$  & $\textbf{0.9999}_{\pm 0.0000}$   & $0.9368_{\pm 0.0000}$   & $0.8588_{\pm 0.0061}$  \\
\multicolumn{1}{c}{}                          & Jin                        & $\textbf{1.0000}_{\pm 0.0000}$           & $0.9864_{\pm 0.0002}$      & $0.9942_{\pm 0.0001}$        & $0.9987_{\pm 0.0000}$         & $0.9921_{\pm 0.0002}$   \\
\multicolumn{1}{c}{}                          & Koza                       & $\textbf{1.0000}_{\pm 0.0000}$           & $0.9965_{\pm 0.0000}$       & $0.9999_{\pm 0.0000}$        & $0.9999_{\pm 0.0000}$      & $0.7571_{\pm 0.0022}$     \\

\midrule
\midrule
Data Group                   & Benchmark                  & $R^2\uparrow$               & $R^2\uparrow$          & $R^2\uparrow$          & $R^2\uparrow$          & $R^2\uparrow$ \\
\cmidrule(lr){1-1} \cmidrule(lr){2-2} \cmidrule(lr){3-3} \cmidrule(lr){4-4} \cmidrule(lr){5-5} \cmidrule(lr){6-6} \cmidrule(lr){7-7}
\multirow{3}{2cm}{\centering SRBench $(d \geq 2)$}& Feynman                & $\textbf{0.9931}_{\pm 0.0015} $         & $0.9806_{\pm 0.0003}$       & $0.9190_{\pm 0.0004}$          & $0.9234_{\pm 0.0000}$      & $0.5641_{\pm 0.0028}$ \\
                                              & Strogatz                   & $\textbf{0.9968}_{\pm 0.0031} $         & $0.9455_{\pm 0.0003}$        & $0.9534_{\pm 0.0003}$         & $0.8816_{\pm 0.0000}$      & $0.6511_{\pm 0.0012}$ \\
                                              & Black-box                  & $\textbf{0.8908}_{\pm 0.0028} $         & $0.6697_{\pm 0.0010}$        & $0.6086_{\pm 0.0021}$         & $0.4226_{\pm 0.0000}$      & $0.4528_{\pm 0.0107}$ \\
\bottomrule
\end{tabular}
}
\vskip -0.1in
\end{table*}

\begin{figure*}[h!]
  \centering
  \vskip 0.1in
  \subfigure[without RC and RC$+$PG]{\includegraphics[width=0.3\textwidth]{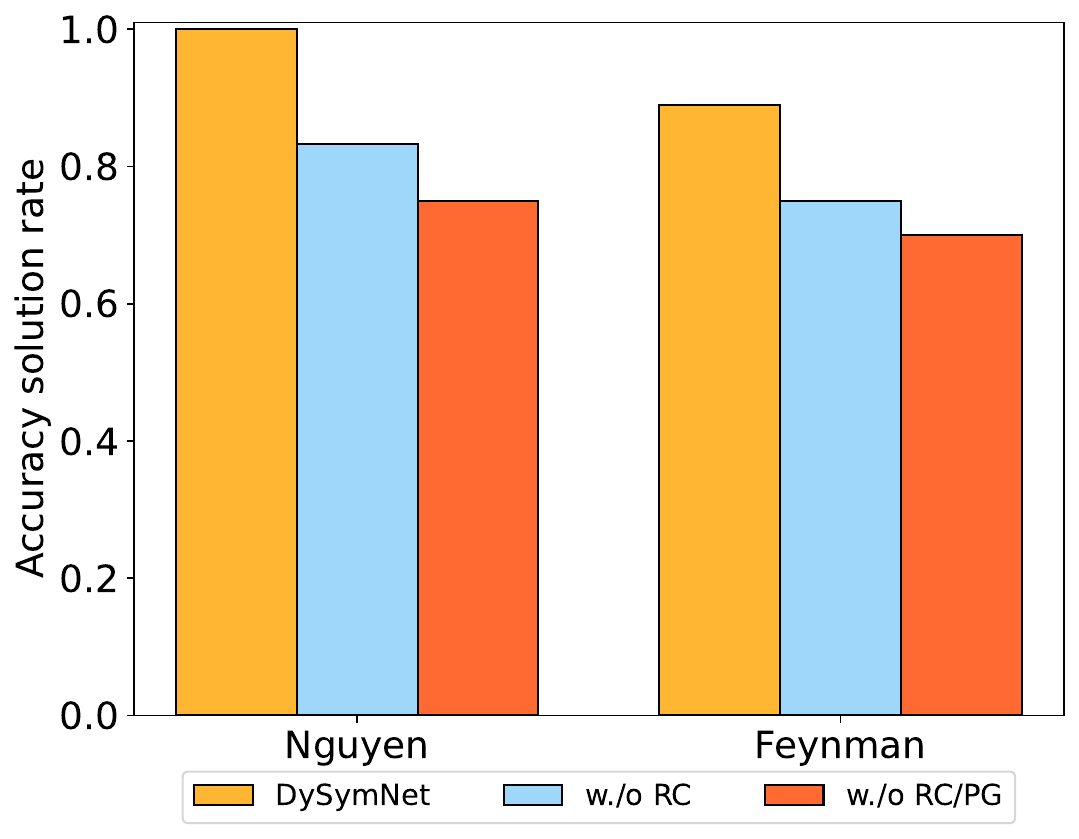}}
  \hfill
  \subfigure[without PG and RC$+$AGC]{\includegraphics[width=0.3\textwidth]{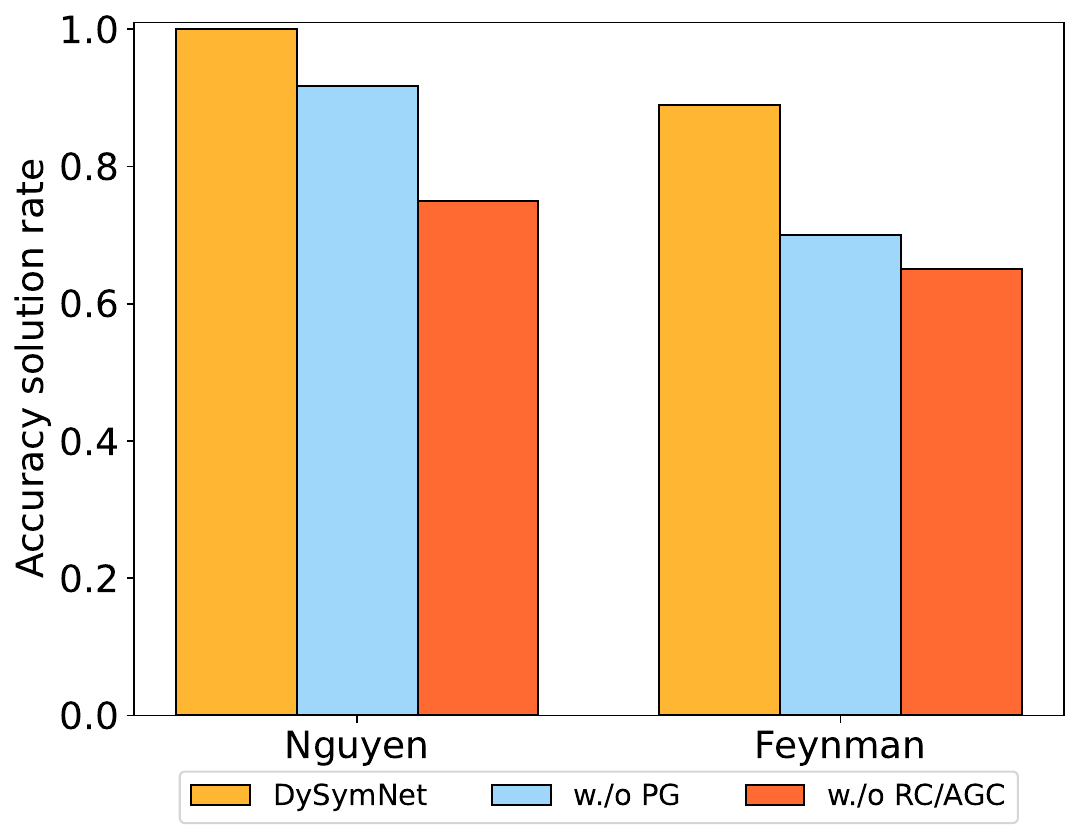}}
  \hfill
  \subfigure[without AGC and PG$+$AGC]{\includegraphics[width=0.3\textwidth]{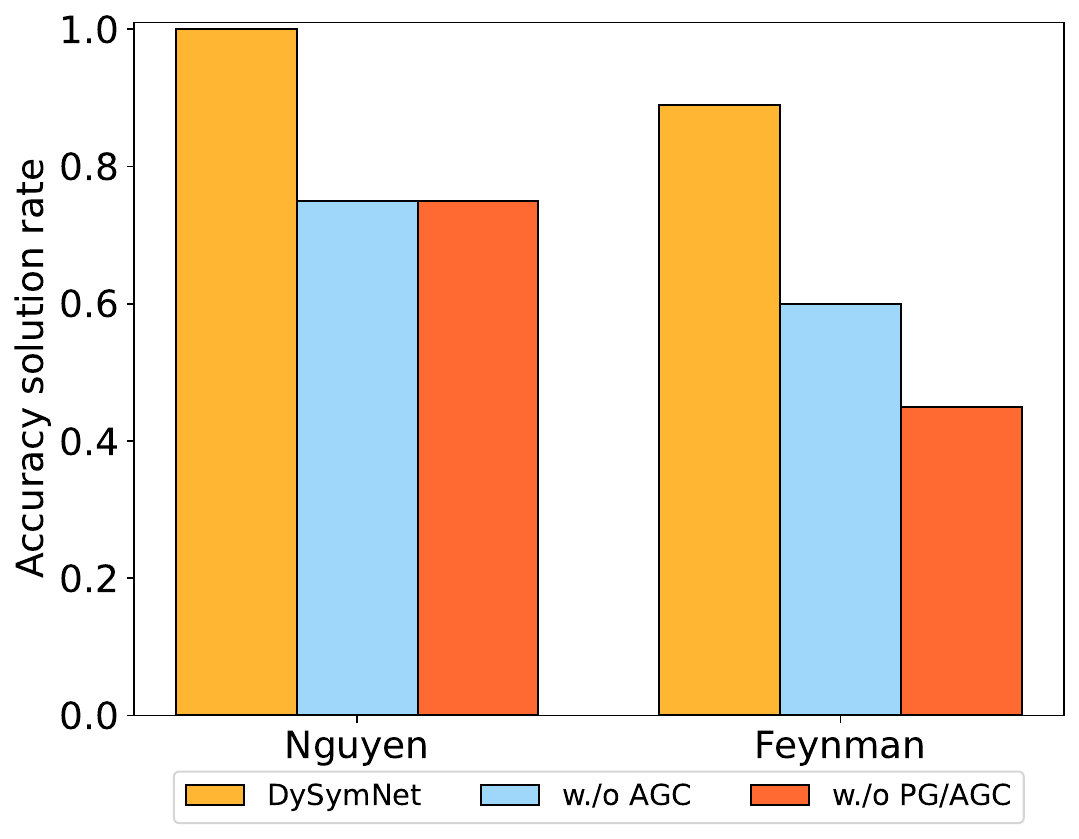}}
  \caption{Ablation study of \dysymnet on Nguyen and Feynman benchmarks. The three subfigures show the performance comparison of Dysymnet without different components, including Refine Constant (\textbf{RC}), Policy Gradient (\textbf{PG}), and Adaptive Gradient Clipping (\textbf{AGC}).}
  \vskip -0.1in
  \label{fig:ablations}
\end{figure*}

\subsection{Baselines}
\label{baselines}
SRBench~\citep{la1contemporary} has reported the performance of 14 SR methods and 7 machine learning methods on \textit{Feynman}, \textit{Black-box}, and \textit{Strogatz} datasets from PMLB~\citep{olson2017pmlb}. We briefly introduce these methods in Appendix~\ref{app:SRBench methods}.
In addition, we compare the performance of our method against four other mainstream SR algorithms currently:
\begin{itemize}[leftmargin=*]
    \item \textbf{A Uniﬁed Framework for Deep Symbolic Regression (uDSR).} A modular, unified framework for symbolic regression that integrates five disparate SR solution strategies~\citep{landajuela2022unified}. 
    %We use the open-source implementation provided by the authors.%\footnote{\url{https://github.com/brendenpetersen/deep-symbolicregression}}
    \item \textbf{Symbolic Regression via Neural-Guided Genetic Programming Population Seeding (NGGP).} A symbolic regression method combining GP and RL~\citep{mundhenk2021symbolic}, the current state-of-the-art algorithm, superseding DSR~\citep{petersen2020deep}. 
    %We use the open-source implementation provided by the authors.%\footnote{\url{https://github.com/brendenpetersen/deep-symbolicregression}}
    \item \textbf{Neural Symbolic Regression that Scales (NeSymReS).} A transformer-based symbolic regression method based on large-scale supervised training~\citep{biggio2021neural}. 
    %We use the open-source implementation provided by the authors.%\footnote{\url{https://github.com/SymposiumOrganization/NeuralSymbolicRegressionThatScales}}
    \item \textbf{Learning Equations for Extrapolation and Control (EQL).} A method of identifying an expression on a pre-set fixed symbolic network~\citep{sahoo2018learning}.
\end{itemize}
All details for baselines are reported in Appendix~\ref{app:baseline details}.

\subsection{Benchmark problem sets}
To facilitate quantitative benchmarking of our and other SR algorithms, we conducted evaluations and comparisons on almost all publicly available benchmark datasets in the SR field. We categorized these datasets into two groups, consisting of eight benchmark datasets. 
% One group comprises four low-dimensional datasets ($d\leq2$) that we named \textbf{Standard benchmarks}. 
\paragraph{\textbf{Standard benchmarks $(d \leq 2)$}}
They are widely used by current SR methods, including the Nguyen benchmark~\citep{uy2011semantically}, Nguyen* (Nguyen with constants)~\citep{petersen2020deep}, Constant, Keijzer~\citep{keijzer2003improving}, Livermore~\citep{mundhenk2021symbolic}, R rationals~\citep{krawiec2013approximating}, Jin~\citep{jin2019bayesian} and Koza~\citep{koza1994genetic}. The complete benchmark functions are given in Appendix~\ref{app:benchmark}. Nguyen was the main benchmark used in~\citep{petersen2020deep}. 
\paragraph{\textbf{SRBench $(d \geq 2)$}}
%The other group is the more challenging \textbf{SRBench} ~\citep{la1contemporary}, 
SRBench~\citep{la1contemporary} is a living SR benchmark that includes datasets with more variables ($d\geq2$). It includes three benchmark datasets: \textit{Feynman}, \textit{Black-box}, and \textit{Strogatz}. The \textit{Feynman} dataset comprises a total of 119 equations sourced from \textit{Feynman Lectures on Physics database} series~\citep{udrescu2020ai}. The \textit{Strogatz} dataset contains 14 SR problems sourced from the \textit{ODE-Strogatz database}~\citep{la2016inference}. The \textit{Black-box} dataset comprises 122 problems without ground-truth expression. The input points $(\bm{x}, y)$ from these three problem sets are provided in Penn Machine Learning Benchmark (PMLB)~\citep{olson2017pmlb} and have been examined in SRBench for the SR task.

\section{Results on Benchmark Problem Sets}

\paragraph{(RQ1) Fitting accuracy on benchmarks}
Table~\ref{tab:fit acc} reports the performance comparison results of our method with the baselines across two group benchmarks. Notably, the problem sets in the standard benchmarks have at most two variables ($d\leq2$) and ground-truth expressions that are relatively simple in form. Our method can quickly find the best expression without requiring many rounds of controller iteration. Thus, our method has a clear advantage in inference efficiency compared to GP-based methods, uDSR, and NGGP, which require iterating through thousands of expressions to find the best one. The results indicate that \dysymnet outperforms the current strong SR baselines in terms of the $R^2$ while maintaining a lower mean squared error (MSE). Additionally, \dysymnet successfully recovered all expressions in the Jin dataset $(d=2)$ and Koza dataset. This highlights the importance of reducing the search space to find optimal expressions.

\paragraph{(RQ2) Ablation studies}
\label{sec:abla}
We performed a series of ablation studies to quantify the effect of each of the components. In figure~\ref{fig:ablations}, we show the accuracy solution rate for \dysymnet on the Nguyen and Feynman benchmarks for each ablation. The results demonstrate the effectiveness of these components in improving performance.

\begin{figure}[htbp]
    \centering
    \vskip 0.1in
    \includegraphics[width=0.9\linewidth]{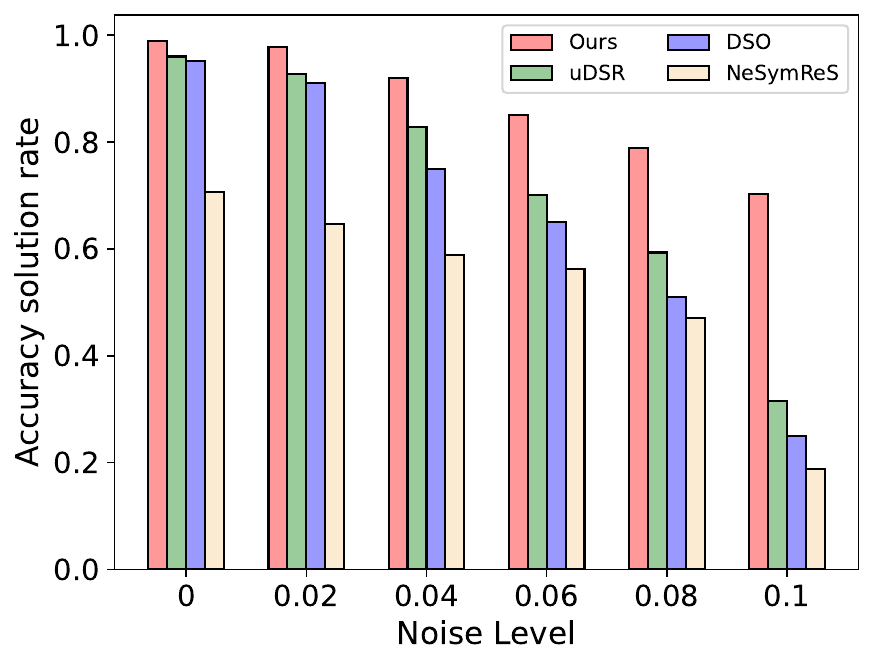}
    \caption{Acuracy solution rate of four approaches on the Standard benchmarks against increasing the noise level.}
    \label{fig:noise}
    \vskip -0.1in
\end{figure}

\paragraph{(RQ3) Performance under noisy data}
Since real data are almost always afflicted with measurement errors or other forms of noise, We investigated the robustness of \dysymnet to noisy data by adding independent Gaussian noise to the dependent variable with mean zero and standard deviation proportional to the root-mean-square of the dependent variable in the training data. In Figure~\ref{fig:noise}, we varied the proportionality constant from $0$ (noiseless) to $10^{-1}$ and evaluated each algorithm across Standard benchmarks. 
\textit{Accuracy solution rate} is defined as the proportion for achieving $R^2 > 0.99$ on the benchmark. The experiments show that our method has the best robustness to noise, and there is no overfitting to noise when a small amount of noise is added, which is significantly better than the NeSymReS based on large-scale supervised training.

\section{(RQ4) Physical Law Discovery: Free Falling Balls with Air Resistance}
It is well known that Galileo discovered the law of free fall of an object by means of two small balls of different masses, which can be formalised as $H(t) = h_0 + v_0 t - \frac{1}{2}gt^2$, where $h_0$ denotes initial height, $v_0$ the initial velocity, and $g$ the gravitational acceleration. and $g$ the gravitational acceleration. However, the above laws of physics apply only to the ideal state and cease to apply when air resistance is encountered. 
Many efforts have been made to uncover the effect of air resistance and to derive mathematical models to describe free-falling objects with air resistance~\citep{clancy1975aerodynamics,lindemuth1971effect,greenwood1986air}.

We performed our experiments on data from 11 different balls, where each ball was experimented with individually.
The measurement dataset of a free-falling ball is divided into a training set (records from the first 2 seconds) and a test set (records after 2 seconds). Three physical models derived from mathematical principles were selected as baseline models for this experiment\footnote{Model-A: \url{https://faraday.physics.utoronto.ca/IYearLab/Intros/FreeFall/FreeFall.html}}$^,$\footnote{Model-B: \url{https://physics.csuchico.edu/kagan/204A/lecturenotes/Section15.pdf}}$^,$\footnote{Model-C: \url{https://en.wikipedia.org/wiki/Free\_fall}}, and the unknown constant coefficient values were estimated using Powell's conjugate direction method. Other methods like uDSR, NGGP, EQL, NeSymRes, and AIFeynman~\citep{udrescu2020ai} are not included as these methods tend to have large generalisation errors due to only a limited number of 20-30 points per training set.
The expressions of three baseline physical models are presented at Table~\ref{tab:models}. 
%as follows: \textbf{Model-A}: $H(t)=c_0 + c_1 t + c_2 t^2 + c_3 t^3$, \textbf{Model-B}: $H(t)=c_0 + c_1 t + c_2 e^{c_3 t}$, and \textbf{Model-C}: $H(t)=c_0 + c_1\log(\cosh(c_2 t))$. 

The physical laws discovered by \dysymnet from training data are evaluated to the test data and compared with the ground truth. Since the $R^2$ obtained by Model-B on the test set are all negative, we use Mean square error (MSE) as the evaluation metric.

\begin{table}[htbp]
        \caption{Expressions of baseline models, where $c_i$ is unknown constants.} 
        \centering
        \vskip 0.1in
        \label{tab:models}
        \begin{tabular}{ll}
        \toprule
        \textbf{Physics Model} & \textbf{Derived model expression}\\
        \midrule
        Model-A  & $H(t)=c_0 + c_1 t + c_2 t^2 + c_3 t^3$\\
        Model-B & $H(t)=c_0 + c_1 t + c_2 e^{c_3 t}$  \\
        Model-C & $H(t)=c_0 + c_1\log(\cosh(c_2 t))$\\
        \bottomrule
        \end{tabular}
        \vskip -0.1in
\end{table}

Table~\ref{tab:ball mse} reported the MSE of \dysymnet and other three physical models. The \dysymnet-discovered physical laws are shown in Appendix Table~\ref{Table:ball eq}, and their predictions in the last 2 seconds of dropping, compared to the ground truth trajectory, are shown in Appendix Figure~\ref{fig:balls res}. It can be concluded that compared to physical laws from human, the physical laws distilled by \dysymnet leads to a better approximation of the free-falling objects with air resistance.

\begin{table}[htbp]
\caption{MSE of the test set. The \dysymnet reaches the best prediction results in most (10 out of 11) cases.}
\label{tab:ball mse}
\vskip 0.1in
\centering
\resizebox{\linewidth}{!}{
\begin{tabular}{ccccc}
\toprule
\textbf{Type} &  \textbf{\dysymnet} &  \textbf{Model-A} & \textbf{Model-B} & \textbf{Model-C} \\
\midrule
baseball             & \textbf{0.0071} & 2.776  & 94.63 & 2.623    \\
blue basketball      & \textbf{0.3713} & 0.5022 & 79.23 & 2.446    \\
green basketball     & 0.2395          & \textbf{0.1029} & 85.44 & 1.242    \\
volleyball           & \textbf{0.2838} & 0.5773 & 80.98 & 0.6238    \\
bowling ball         & \textbf{0.0020} & 0.3282 & 87.06 & 2.471    \\
golf ball            & \textbf{0.0380} & 0.2117 & 86.12 & 1.005    \\
tennis ball          & \textbf{0.0052} & 0.2523 & 72.28 & 0.06219    \\
whiffle ball 1       & \textbf{0.0055} & 1.883  & 65.40 & 0.2410    \\
whiffle ball 2       & \textbf{0.4923} & 0.6223 & 58.49 & 0.9178    \\
yellow whiffle ball  & \textbf{0.4558} & 17.33 & 44.99  & 2.544    \\
orange whiffle ball  & \textbf{0.0090} & 0.3827 & 36.75 & 3.317    \\
\midrule
\textbf{Average}     & \textbf{0.1736} & 2.2698 & 71.94 & 1.5902    \\
\bottomrule
\end{tabular}
}
\vskip -0.1in
\end{table}

\section{Conclusion}
In this work, we propose \dysymnet, a neural-guided dynamic symbolic network for SR. We provide a new search paradigm for SR that automatically identify the best expression from different symbolic network guided by RL. 
Through extensive numerical experiments, we demonstrate the effectiveness of \dysymnet in solving high-dimensional problems. We also show the value of the application of \dysymnet to real-world problems, where the physical laws explored by \dysymnet provide a better approximation of the falling ball compared to three physical models in the falling ball experiment with air resistance. Future research may focus on enhancing the pruning techniques of symbolic networks to improve the comprehensive performance and inference efficiency of the model.

\section*{Acknowledgements}

This work was supported by the National Natural Science Foundation of China (No. 92370117), and CAS Project for Young Scientists in Basic Research (No. YSBR-090).

\section*{Impact Statement}
The research presented in this paper is aimed at advancing the field of Machine Learning. After careful consideration, we believe that although any technological advancement has the potential to impact society at large, our work does not entail ethical issues or foresee any potential harm to society. Therefore, we do not specifically highlight any societal consequences here. Should our research be applied in real-world scenarios in the future, we will ensure that it is conducted within a framework of strict ethical standards and social responsibility.


\bibliographystyle{icml2024}
\renewcommand{\bibname}{References}
\bibliography{example_paper}

%%%%%%%%%%%%%%%%%%%%%%%%%%%%%%%%%%%%%%%%%%%%%%%%%%%%%%%%%%%%%%%%%%%%%%%%%%%%%%%
%%%%%%%%%%%%%%%%%%%%%%%%%%%%%%%%%%%%%%%%%%%%%%%%%%%%%%%%%%%%%%%%%%%%%%%%%%%%%%%
% APPENDIX
%%%%%%%%%%%%%%%%%%%%%%%%%%%%%%%%%%%%%%%%%%%%%%%%%%%%%%%%%%%%%%%%%%%%%%%%%%%%%%%
%%%%%%%%%%%%%%%%%%%%%%%%%%%%%%%%%%%%%%%%%%%%%%%%%%%%%%%%%%%%%%%%%%%%%%%%%%%%%%%
\newpage
\appendix
\onecolumn

\section*{\textbf Appendix for ``A Neural-Guided Dynamic Symbolic Network for
Exploring Mathematical Expressions from Data''}

\section*{Contents of Appendix:}
\subsection*{Method details:}
\begin{itemize}
    \item Appendix~\ref{app:code}: Pseudocode for \dysymnet
\end{itemize}

\subsection*{Experimental details:}
\begin{itemize}
    \item Appendix~\ref{app:details}: Computing Infrastructure and Hyperparameter Settings
    \item Appendix~\ref{app:baseline details}: Baselines Algorithms Details
    \item Appendix~\ref{app:SRBench methods}: Short Descriptions of SRBench Baselines
    \item Appendix~\ref{app:benchmark}: Standard Benchmark Expressions
\end{itemize}

\subsection*{Results:}
\begin{itemize}
    \item Appendix~\ref{app:extra}: Extrapolation Experiment
    \item Appendix~\ref{app:conver}: Convergence Analysis
    \item Appendix~\ref{app:balls2}: Free falling Balls with Air Resistance
\end{itemize}

\newpage

\section{Pseudocode for \dysymnet}
\label{app:code}
In this section, we provide the pseudocode for \dysymnet. The overall algorithm is detailed in Algorithm~\ref{alg:dysymnet}. Moreover, we provide pseudocode for the function \texttt{SampleSymbolicNetwork} (line 4 of Algorithm 1) and \texttt{TrainSymbolicNetwork} (line 5 of Algorithm 1) in Algorithm~\ref{alg:sample} and Algorithm~\ref{alg:train}, respectively.
Specifically, Algorithm~\ref{alg:dysymnet} describes the overall framework of \dysymnet. We use RNN as the controller to sample a batch of symbolic network structures and instantiate them. By training the symbolic network, we can obtain compact expressions. Then, we refine the constants using BFGS and calculate corresponding rewards. When we obtain a batch of rewards, we calculate the $\epsilon$-quantile of the rewards. We use the top $(1-\epsilon)$ rewards to calculate policy gradients and corresponding entropies to update the RNN. This iterative process continues until we reach the threshold we set.

\begin{algorithm}[h]
\caption{A neural-guided dynamic symbolic network for symbolic regression}
\label{alg:dysymnet}
\begin{algorithmic}[1]
\INPUT learning rate $\alpha_c$ of controller; entropy coefficient $\lambda_\mathcal{H}$; risk factor $\epsilon$; batch size $N$; sample epochs $M$; reward function $R$; early stop threshold $R_t$; SR problem $\mathcal{P}$ consisting of tabular data $\mathcal{D}(X, y)$
% \KwIn{Symbolic regression problem $\mathcal{P}$ consisting of tabular data $\mathcal{D}(X, y)$}
\OUTPUT Best fitting expression $\tilde{f}^*$
% \SetKwFunction{ShiftRight}{ShiftRight}
\STATE Initialize controller RNN with parameters $\theta_c$, defining distribution over symbolic network $p(\mathcal{S}_{net} | \theta_c)$
% \BlankLine
\FOR{$i \gets 1$ {\bfseries to} $M$}
    \FOR{$j \gets 1$ {\bfseries to} $N$}
        \STATE $\mathcal{S}_{net} \leftarrow \textrm{SampleSymbolicNetwork}(\theta_c)$ \RCOMMENT{Sample a symbolic network via controller}
        \STATE $\tilde{f}_j \leftarrow \textrm{TrainSymbolicNetwork}(\mathcal{S}_{net}, \mathcal{D})$ \RCOMMENT{Train the symbolic network}
        \STATE $\tilde{f}_j^*, \textrm{MSE}_j \leftarrow \textrm{BFGS}(\tilde{f}_j, \mathcal{D})$ \RCOMMENT{Refine constants using BFGS}
        \STATE $R(\tilde{f}_j^*) \leftarrow 1/(1+\textrm{MSE}_j)$ \RCOMMENT{Compute reward through MSE} 
        \IF{$R(\tilde{f}_j^*) > R_t$}
        \STATE \textbf{return} $\tilde{f}_j^*$ \RCOMMENT{Return the best expression if reach the threshold}
        \ENDIF
    \ENDFOR 
    % \begin{ALC@rpt}
    \STATE $\mathcal{R} \leftarrow \{R(\tilde{f}_j^*) \}_{j=1}^N$ \RCOMMENT{Compute batch rewards}
    \STATE $R_\epsilon \leftarrow (1-\epsilon)$-quantile of $\mathcal{R}$ \RCOMMENT{Compute reward threshold}
    \STATE $\mathcal{F} \leftarrow \{\tilde{f}_j^*: R(\tilde{f}_j^*) \geq R_\epsilon\}$ \RCOMMENT{Select subset of expressions above threshold}
    \STATE $\mathcal{R} \leftarrow \{R(\tilde{f}_j^*): R(\tilde{f}_j^*) \geq R_\epsilon \}$ \RCOMMENT{Select corresponding subset of rewards}
    \STATE $\hat{g_1} \leftarrow \textrm{ReduceMean}( (\mathcal{R} - R_\eps) \nabla_{\theta_c} \log p(\mathcal{F} | \theta_c) )$ \RCOMMENT{Compute risk-seeking policy gradient}
    \STATE $\hat{g_2} \leftarrow \textrm{ReduceMean}(\lambda_{\mathcal{H}} \nabla_{\theta_c} \mathcal{H}(\mathcal{F} | \theta_c)) $ \RCOMMENT{Compute risk-seeking policy gradient}
    \STATE $\theta_c \leftarrow \theta_c + \alpha (\hat{g_1} + \hat{g_2})$ \RCOMMENT{Apply gradients}
    % \end{ALC@rpt}
\ENDFOR
\end{algorithmic}
\end{algorithm}

Algorithm~\ref{alg:sample} describes the specific process of using RNN to sample a symbolic network structure, which is corresponding to \textit{Generative process} of Section~\ref{sec:gen}. In a particular sampling process, we first sample from the number library of layers to determine the current number of layers in the symbolic network. Then, we iterate through each layer to decide the number of operators and operator categories for each layer. Once a complete symbolic network structure has been sampled, we instantiate it. The various libraries of symbolic network structure are reported in Table~\ref{tab:hyper}.
\begin{algorithm}[htbp]
    \caption{Sample a symbolic network autoregressively via controller RNN}
    \label{alg:sample}
\begin{algorithmic}[1]
    \INPUT RNN with parameters $\theta_c$; number library for symbolic network layers $\mathcal{L}_L$; number library for operators in each layer $\mathcal{L}_n$; operators library $\mathcal{L}_o$ 
    \OUTPUT Symbolic network $\mathcal{S}_{net}$
    % \BlankLine
    \STATE $c_0 \leftarrow \vec{0}$ \RCOMMENT{Initialize RNN cell state to zero} \\
    \STATE $x \leftarrow \textrm{rand}()$ \RCOMMENT{Initialize input randomly} \\
    \STATE $(\psi_L^{(1)}, c_1) \leftarrow$ \textrm{RNN\_Cell}$(x, c_{0};\theta_c)$ \RCOMMENT{Emit probabilities over $\mathcal{L}_L$}; update state  \\
    %$\psi^{(i)} \leftarrow \textrm{Layer\_Decoder}(\psi^{(i)})$ \RCOMMENT{}\\
    \STATE $L \leftarrow \textrm{Categorical}(\psi_L^{(1)})$ \RCOMMENT{Sample a value from $\mathcal{L}_L$} as the number of layers\\
    \STATE $x \leftarrow \psi_L^{(1)}$ \RCOMMENT{Take the probabilities over $\mathcal{L}_L$ as input}\\
    \FOR{$i \gets 1$ {\bfseries to} $L$}
        \STATE $(\psi_n^{(i)}, c_i) \leftarrow$ \textrm{RNN\_Cell}$(x, c_{i-1};\theta_c)$ \RCOMMENT{Emit probabilities over $\mathcal{L}_n$}; update state  \\
        \STATE $n^{(i)} = \textrm{Categorical}(\psi_n^{(i)})$ \RCOMMENT{Sample a value from $\mathcal{L}_n$ as the number of operators for layer $L^i$}\\
        \STATE $x \leftarrow \psi_n^{(i)}$ \RCOMMENT{Take the probabilities over $\mathcal{L}_n$ as input}\\
        \STATE $(\psi_o^{(i)}, c_i) \leftarrow$ \textrm{RNN\_Cell}$(x, c_{i-1};\theta_c)$ \RCOMMENT{Emit probabilities over $\mathcal{L}_o$}; update state  \\
        \STATE $o^{(i)} \leftarrow  \textrm{Categorical}(\psi_o^{(i)})$ \RCOMMENT{Sample $n^{i}$ operators for layer $L^i$}\\
        \STATE $x \leftarrow \psi_o^{(i)}$ \RCOMMENT{Take the probabilities over $\mathcal{L}_o$ as input}\\    
    \ENDFOR
    \STATE $\mathcal{S}_{net} \leftarrow \textrm{Instantiation}(L, n, o)$ \RCOMMENT{Instantiate a corresponding symbolic network}\\
    \STATE \textbf{return} $\mathcal{S}_{net}$
\end{algorithmic}
\end{algorithm}

Algorithm~\ref{alg:train} describes the specific process of training a symbolic network, which is corresponding to Section~\ref{sub:identify}. We divide the training process into two stages. In the first stage, we use MSE loss to supervise and converge the weights of the symbolic network to an appropriate interval. In the second stage, we add $L_{0.5}^*$ regularization term to sparse the symbolic network and finally perform pruning to obtain a compact expression. We use the adaptive gradient clipping technology described in Section~\ref{sub:identify} in both stages, which makes the training process of the symbolic network more stable.
\begin{algorithm}[htbp]
    \caption{Training process of symbolic network}
    \label{alg:train}
\begin{algorithmic}[1]
    \INPUT learning rate $\alpha_s$, regulation weight $\lambda$; training epochs $n_1$; training epochs $n_2$; window size $w$; weight clipping factor $c$; clipping threshold $\gamma$; pruning threshold $\beta$
    \OUTPUT expression $\tilde{f}$
    % \BlankLine
    \STATE Initialize symbolic network with parameters $\bm{\Theta}$\\
    \STATE $\textrm{queue} \leftarrow []$ \RCOMMENT{Initialize gradient norm queue}\\
    \FOR{$i \gets 1$ {\bfseries to} $n_1$}
        \STATE $J(\bm{\Theta}) \leftarrow \textrm{MSE}(y, \tilde{y})$ \RCOMMENT{Compute MSE loss}\\
        \STATE $\textrm{queue}.\textrm{append}(\Vert \bm{\Theta} \Vert_2)$ \RCOMMENT{Append current gradient norm to queue}\\
        \IF{\upshape len(queue) $>$ $w$}
            \STATE queue.popleft()
        \ENDIF
        \STATE $\gamma \leftarrow \frac{c}{w}\sum_{i=1}^w \sum_{\ell=1}^L \Vert \bm{\Theta}^{(\ell)} \Vert_2$ \RCOMMENT{Compute gradient clipping threshold}\\
        \STATE $g \leftarrow \textrm{clip\_grad\_norm}(\bm{\Theta}, \gamma)$ \RCOMMENT{Clip gradient}\\
        \STATE $\bm{\Theta} \leftarrow \bm{\Theta} - \alpha_s \nabla J(\bm{\Theta})$ \RCOMMENT{Apply gradient}
    \ENDFOR
    \FOR{$i \gets 1$ {\bfseries to} $n_2$}
        \STATE $J(\bm{\Theta}) \leftarrow \textrm{MSE}(y, \tilde{y}) + \lambda L_{0.5}^*(\bm{\Theta})$ \RCOMMENT{Compute MSE loss and regularizaiton loss}\\
        \STATE $\textrm{queue}.\textrm{append}(\Vert \bm{\Theta} \Vert_2)$ \RCOMMENT{Append current gradient norm to queue}\\
        \IF{\upshape len(queue) $>$ $w$}
            \STATE queue.popleft()
        \ENDIF
        \STATE $\gamma \leftarrow \frac{c}{w}\sum_{i=1}^w \sum_{\ell=1}^L \Vert \bm{\Theta}^{(\ell)} \Vert_2$ \RCOMMENT{Compute gradient clipping threshold}\\
        \STATE $g \leftarrow \textrm{clip\_grad\_norm}(\bm{\Theta}, \gamma)$ \RCOMMENT{Clip gradient}\\
        \STATE $\bm{\Theta} \leftarrow \bm{\Theta} - \alpha_s \nabla J(\Theta)$ \RCOMMENT{Apply gradient}
    \ENDFOR
    \STATE $\tilde{f} \leftarrow \textrm{Pruning}(\bm{\Theta}, \beta)$ \RCOMMENT{Pruning and using Sympy to convert to corresponding expression}\\
    \STATE \textbf{return} $\tilde{f}$ 
\end{algorithmic}
\end{algorithm}

\section{Computing Infrastructure and Hyperparameter Settings}
\label{app:details}
In this section, we describe additional experimental details, including the experimental environment, hyperparameter settings of our approach and other baselines. 
\paragraph{Computing infrastructure}
Experiments in this work were executed on an Intel Xeon Platinum 8255C CPU @ 2.50GHz, 32GB RAM equipped with NVIDIA Tesla V100 GPUs 32 GB.

\paragraph{Hyperparameter settings}
The hyperparameters of our method mainly include three parts, which are the hyperparameters of controller RNN, the hyperparameters of symbolic network training, and the hyperparameters of symbolic network structure. For the RNN, the space of hyperparameters considered was learning rate $\in \{0.0003, 0.0006, 0.001\}$, and entropy weight $\lambda_\mathcal{H} \in \{0.001, 0.005, 0.01\}$. For the symbolic network training, the space of hyperparameters considered was learning rate $\in \{0.01, 0.05, 0.1\}$, weight pruning threshold $\beta \in \{0.001, 0.01, 0.05\}$, adaptive gradient clipping $\in \{\textsc{True}, \textsc{False}\}$, window size $w \in \{10, 50, 100\}$, and add bias $\in \{\textsc{True}, \textsc{False}\}$. Adding bias can further improve the performance of the algorithm, but it also increases the complexity of the final expression. We tuned hyperparameters by performing grid search on benchmarks Nguyen-5 and Nguyen-10. We selected the hyperparameters combination with the highest average $R^2$. The best found hyperparameters are listed in Table~\ref{tab:hyper} and used for all experiments and all benchmark expressions.
For the symbolic network structure, there are three categories of hyperparameters that are the number of layers, the number of operators for each layer, and the type of each operator. Notably, these parameters differ from the model training parameters. Each structural parameter is a set that RNN can choose from during sampling, and there is no optimal value for them, as they only determine the upper limit of the symbolic network's representational capacity, similar to the maximum sequence length defined in NGGP~\citep{mundhenk2021symbolic}. Thus, the selection of these parameters can be adjusted based on specific scenarios to achieve the best results. For fair comparison, we used the same structural hyperparameter in all experiments, where the operator library was consistent with that of uDSR and NGGP.

\begin{table}[htbp]
\centering
\caption{Hyperparameters for \dysymnet.}
\label{tab:hyper}
\vskip 0.1in
\begin{tabular}{ccc}
\toprule
\multicolumn{1}{l}{\textbf{Hyperparameter}} & \textbf{Symbol} & \textbf{Value}   \\
\midrule
\multicolumn{1}{l}{\textbf{RNN Parameters}}   \\
\midrule
\multicolumn{1}{l}{Learning rate}  & $\alpha_c$  &  0.0006 \\
\multicolumn{1}{l}{Entropy weight} &  $\lambda_\mathcal{H}$  &  0.005 \\
\multicolumn{1}{l}{RNN cell size} & -- & 32 \\
\multicolumn{1}{l}{RNN cell layers} & -- & 1 \\
\multicolumn{1}{l}{Risk factor} & $\epsilon$ & 0.5 \\

\midrule
\multicolumn{1}{l}{\textbf{Symbolic network training parameters}} \\
\midrule
\multicolumn{1}{l}{Learning rate} & $\alpha_s$  & 0.1 \\
\multicolumn{1}{l}{Regularization weight} & $\lambda$ & 0.005 \\
\multicolumn{1}{l}{Weight pruning threshold} & $\beta$ & 0.01 \\
\multicolumn{1}{l}{Training epochs (stage 1)} & $n_1$ & 10000 \\
\multicolumn{1}{l}{Training epochs (stage 2)} & $n_2$ & 10000 \\
\multicolumn{1}{l}{Adaptive gradient clipping} & -- & $\textsc{True}$ \\
\multicolumn{1}{l}{Window size} & $w$ & 50 \\
\multicolumn{1}{l}{Add bias} & -- & $\textsc{False}$\\
\midrule
\multicolumn{1}{l}{\textbf{Symbolic network structure parameters}} \\
\midrule
\multicolumn{1}{l}{Operators library}  & -- & $\{+, -, \times, \sin, \cos, \tan, \exp, \log, \cosh, \placeholder^2\}$ \\
\multicolumn{1}{l}{Number library of layers} & -- & $\{1, 2, 3, 4, 5\}$ \\
\multicolumn{1}{l}{Number library of operators for each layer} & -- & $\{1, 2, 3, 4, 5, 6\}$ \\
\bottomrule
\end{tabular}
\vskip -0.1in
\end{table}

\section{Baselines Algorithms Details}
\label{app:baseline details}
\paragraph{A Unified Framework for Deep Symbolic Regression (uDSR)} \citet{landajuela2022unified} proposed a modular, unified framework for SR that integrates five disparate SR solution strategies. In this work, we use the open-source code and implementation provided by the authors. To ensure fair comparison, we use the same symbol library as our algorithm and add the $\textsc{const}$ token to give it the ability to generate constants. In addition, we used the hyperparameter combination for training reported in the literature.

\paragraph{Deep Symbolic Optimization (NGGP)}
NGGP~\citep{mundhenk2021symbolic} is a hybrid method that combines RL and GP. They use a policy network to seed the starting population of a GP algorithm. In this work, we use the open-source code and implementation provided by the authors\footnote{\url{https://github.com/brendenpetersen/deep-symbolicregression}}. To ensure fair comparison, we use the same symbol library as our algorithm and add the $\textsc{const}$ token to give it the ability to generate constants. In addition, we used the hyperparameter combination for training reported in the literature.

\paragraph{Neural Symbolic Regression that Scales (NeSymReS)}
NeSymReS~\citep{biggio2021neural} is a transformer-based model for SR. It is trained on synthetic data with a scale of 10M and 100M. In this work, we use the open-source code and implementation provided by the authors\footnote{\url{https://github.com/SymposiumOrganization/NeuralSymbolicRegressionThatScales}}, following their proposal, and using their released 100M model. To ensure fair comparison, we use the settings reported in the literature, with a beam size of 32 and 4 restarts of BFGS per expression. 
Due to the limitation of training data, NeSymReS only supports to inference the problem with up to three variables. Following~\citep{landajuela2022unified}, we utilize the sklearn library~\citep{pedregosa2011scikit} to select the top-$k$ $(k=3)$ most relevant features when inferring expressions with more than three variables on SRBench.

\paragraph{Learning Equations for Extrapolation and Control (EQL)}
EQL~\citep{sahoo2018learning} is a fully-connected network where elementary functions are used as activation functions. Whenever reasoning about a new problem, the user needs to first determine the structure of the network. In this work, we use the open-source code and implementation provided by the authors\footnote{\url{https://github.com/martius-lab/EQL}}.

\section{Short Descriptions of SRBench Baselines}
\label{app:SRBench methods}
Herein, we present concise descriptions of the 14 SR baseline methodologies employed by SRBench, as depicted in Figure~\ref{fig:perato}. These 14 methods include both GP-based and deep learning-based approaches and do not encompass seven other well-known machine learning methods.
\begin{itemize}
    \item \textbf{Bayesian symbolic regression (BSR):} \citet{jin2019bayesian} present a method that incorporates prior knowledge and generates expression trees from a posterior distribution by utilizing a highly efficient Markov Chain Monte Carlo algorithm. This approach is impervious to hyperparameter settings and yields more succinct solutions as compared to solely GP-based methods.

    \item \textbf{AIFeynman:} AIFeynman~\citep{udrescu2020ai} exploits knowledge of physics and the given training data, by identifying simplifying properties (e.g., multiplicative separability) of the underlying functional form. They decompose a larger problem into several smaller sub-problems and resolve each sub-problem through brute-force search.

    \item \textbf{Age-fitness Pareto optimization (AFP):}
    \citet{schmidt2010age} propose a genetic programming (GP) approach that mitigates premature convergence by incorporating a multidimensional optimization objective that assesses solutions based on both their fitness and age.

    \item \textbf{AFP with co-evolved fitness estimates (AFP-FE):}
    \citet{schmidt2009distilling} expands on ~\citep{schmidt2008coevolution} by adding a new fitness estimation approach.

    \item $\epsilon$\textbf{-lexicase selection (EPLEX):}
    \citet{la2019probabilistic} present an approach that enhances the parent selection procedure in GP by rewarding expressions that perform well on more challenging aspects of the problem rather than assessing performance on data samples in aggregate or average.

    \item \textbf{Feature engineering automation tool (FEAT):}
    \citet{la2018learning} present a strategy for discovering simple solutions that generalize well by storing solutions with accuracy-complexity trade-offs to increase generalization and interpretation.

    \item \textbf{Fast function extraction (FFX):}
    \citet{mcconaghy2011ffx} propose a non-evolutionary SR technique based on pathwise learning~\citep{friedman2010regularization} that produces a set of solutions that trade-off error versus complexity while being orders of magnitude faster than GP and giving deterministic convergence.

    \item \textbf{GP (gplearn):}
    The open source library \texttt{gplearn} (\url{https://github.com/trevorstephens/gplearn}) is used in Koza-style GP algorithms. This implementation is very similar to the GP component in NGGP and uDSR.

    \item \textbf{GP version of the gene-pool optimal mixing evolutionary algorithm (GP-GOMEA):}
    \citet{virgolin2021improving} propose combining GP with linkage learning, a method that develops a model of interdependencies to predict which patterns will propagate and proposes simple and interpretable solutions.

    \item \textbf{Deep symbolic regression (DSR):}
    \citet{petersen2020deep} leverage a recurrent neural network to sample the pre-order traversal of the expression guiede by the proposed risk-seeking gradient. 

    \item \textbf{Interaction-transformation evolutionary algorithm (ITEA):}
    \citet{de2021interaction} present a mutation-based evolutionary algorithm based on six mutation heuristics that allows for the learning of high-performing solutions as well as the extraction of the significance of each original feature of a data set as an analytical function.
    
    \item \textbf{Multiple regression genetic programming (MRGP):}
    As a cost-neutral modification to basic GP, \citet{arnaldo2014multiple} describe an approach that decouples and linearly combines sub-expressions via multiple regression on the target variable.

    \item \textbf{SR with Non-Linear least squares (Operon):}
    \citet{kommenda2020parameter} improve generalization by including nonlinear least squares optimization into GP as a local search mechanism for offspring selection.

    \item \textbf{Semantic backpropagation genetic programming (SBP-GP):}
    \citet{virgolin2019linear} improve the random desired operator algorithm~\citep{pawlak2014semantic}, a semantic backpropagation-based GP approach, by introducing linear scaling concepts, making the process significantly more effective despite being computationally more expensive.
    
\end{itemize}

\section{Standard Benchmark Expressions}
\label{app:benchmark}
This section describes the exact expressions in the standard benchmarks ($d \leq 2$) used to compare our method with the current mainstream baselines. In table~\ref{tab:benchmarks}, we show the name of the benchmark, corresponding expressions, and dataset information. For fair comparison, the dataset used in all methods was generated under the same seed.

\begin{longtable}[htbp]{lll}
\caption{Standard benchmark symbolic regression problem specifications. Input variables are denoted by $x$ and/or $y$. $U\left(a, b, c\right)$ denotes $c$ random points uniformly sampled between $a$ and $b$ for each input variable; training and test datasets use different random seeds.}
\label{tab:benchmarks}
    \\ \toprule
    Name  & Expression & Dataset \\
    \midrule
    Nguyen-1 & $x^3+x^2+x$ & $U\left(-1, 1, 256\right)$ \\
    Nguyen-2 & $x^4+x^3+x^2+x$ & $U\left(-1, 1, 256\right)$ \\
    Nguyen-3 & $x^5+x^4+x^3+x^2+x$ & $U\left(-1, 1, 256\right)$ \\
    Nguyen-4 & $x^6+x^5+x^4+x^3+x^2+x$ & $U\left(-1, 1, 256\right)$ \\
    Nguyen-5 & $\sin(x^2)\cos(x)-1$ & $U\left(-3, 3, 256\right)$ \\
    Nguyen-6 & $\sin(x)+\sin(x+x^2)$ & $U\left(-3, 3, 256\right)$ \\
    Nguyen-7 & $\log(x+1)+\log(x^2+1)$ & $U\left(0, 2, 256\right)$ \\
    Nguyen-8 & $\sqrt{x}$ & $U\left(0, 4, 256\right)$ \\
    Nguyen-9 & $\sin(x)+\sin(y^2)$ & $U\left(0, 1, 256\right)$ \\
    Nguyen-10 & $2\sin(x)\cos(y)$ & $U\left(0, 1, 256\right)$ \\
    Nguyen-11 & $x^y$ & $U\left(0, 1, 256\right)$ \\
    Nguyen-12 & $x^4-x^3+\frac{1}{2}y^2-y$ & $U\left(0, 1, 256\right)$ \\
    \midrule
    Nguyen-1\textsuperscript{c} & $3.39x^3+2.12x^2+1.78x$ & $U\left(-1, 1, 256\right)$ \\
    Nguyen-5\textsuperscript{c} & $\sin(x^2)\cos(x)-0.75$ & $U\left(-1, 1, 256\right)$ \\
    Nguyen-7\textsuperscript{c} & $\log(x+1.4)+\log(x^2+1.3)$ & $U\left(0, 2, 256\right)$ \\
    Nguyen-8\textsuperscript{c} & $\sqrt{1.23 x}$ & $U\left(0, 4, 256\right)$ \\
    Nguyen-10\textsuperscript{c} & $\sin(1.5x)\cos(0.5y)$ & $U\left(0, 1, 256\right)$ \\
    \midrule
    Constant-1 & $3.39x^3+2.12x^2+1.78x$ & $ U\left(-1, 1, 256\right)$ \\ 
    Constant-2 & $\sin{(x^2)} \cos{(x)}-0.75$ & $U\left(-1, 1, 256\right)$\\
    Constant-3 & $\sin{(1.5x)} \cos{(0.5y)}$ & $U\left(0, 1, 256\right)$ \\ 
    Constant-4 & $2.7x^y$ & $U\left(0, 1, 256\right)$ \\ 
    Constant-5 & $\sqrt{1.23x}$ & $U\left(0, 4, 256\right)$ \\ 
    Constant-6 & $x^{0.426}$  & $U\left(0, 2, 256\right)$\\ 
    Constant-7 & $2\sin{(1.3x)} \cos{(y)}$ & $U\left(-1, 1, 256\right)$\\ 
    Constant-8 & $\log(x+1.4)+\log(x^2+1.3)$ & $U\left(0, 2, 256\right)$ \\ 
    \midrule
    Keijzer-3 & $0.3 \cdot x \cdot \sin{(2 \cdot \pi \cdot x)}$ & $U\left(-1, 1, 256\right)$ \\ 
    Keijzer-4 & $x^3\exp{(-x)}\cos{(x)} \sin{(x)} ({\sin{x}}^2\cos{x}-1)$ & $U\left(-1, 1, 256\right)$\\ 
    Keijzer-6 & $\frac{x\cdot(x+1)}{2}$ & $U\left(-1, 1, 256\right)$ \\ 
    Keijzer-7 & $\ln{x}$ & $U\left(1, 2, 256\right)$ \\ 
    Keijzer-8 & $\sqrt{x}$ & $U\left(0, 1, 256\right)$ \\ 
    Keijzer-9 & $\log{(x+\sqrt{(x^2+1)})}$ & $U\left(-1, 1, 256\right)$ \\ 
    Keijzer-10 & $x^y$  & $U\left(0, 1, 256\right)$ \\ 
    Keijzer-11 & $xy+\sin{((x-1)(y-1))}$  & $U\left(-1, 1, 256\right)$ \\ 
    Keijzer-12 & $x^4-x^3+\frac{y^2}{2}-y$  & $U\left(-1, 1, 256\right)$ \\ 
    Keijzer-13 & $6\sin{(x)} \cos{(y)}$ & $U\left(-1, 1, 256\right)$\\ 
    Keijzer-14 & $\frac{8}{2+x^2+y^2}$ & $U\left(-1, 1, 256\right)$\\ 
    Keijzer-15 & $\frac{x^3}{5}+\frac{y^3}{2}-y-x$  & $U\left(-1, 1, 256\right)$\\ 
    \midrule
    Livermore-1 & $\frac{1}{3}+x+\sin\left({x}^2\right)$ & $U\left(-5, 5, 256\right)$ \\
    Livermore-2 & $\sin\left({x}^2\right) \cos\left(x\right)-2$ & $U\left(-1, 1, 256\right)$ \\
    Livermore-3 & $\sin\left({x}^3\right) \cos\left({x}^2\right)-1$ & $U\left(-1,1,256\right)$ \\
    Livermore-4 & $\log(x+1)+\log({x}^2+1)+\log(x)$ & $U\left(0,2,256\right)$ \\
    Livermore-5 & ${x}^4-{x}^3+{x}^2-y$ & $U\left(-1,1,256\right)$ \\
    Livermore-6 & $4{x}^4+3{x}^3+2{x}^2+x$ & $U\left(-1,1,256\right)$ \\
    Livermore-7 & $\sinh(x)$ & $U\left(-1,1,256\right)$ \\
    Livermore-8 & $\cosh(x)$ & $U\left(-1,1,256\right)$ \\
    Livermore-9 & ${x}^9+{x}^8+{x}^7+{x}^6+{x}^5+{x}^4+{x}^3+{x}^2+x$ & $U\left(-1,1,256\right)$ \\
    Livermore-10 & $6\sin\left(x\right) \cos\left(y\right)$ & $U\left(-1,1,256\right)$ \\
    Livermore-11 & $\frac{{x}^2 {x}^2}{x+y}$ & $U\left(-1,1,256\right)$ \\
    Livermore-12 & $\frac{{x}^5}{{y}^3}$ & $U\left(-1,1,256\right)$ \\
    Livermore-13 & ${x}^{\frac{1}{3}}$ & $U\left(0,1,256\right)$ \\
    Livermore-14 & ${x}^3+{x}^2+x+\sin\left(x\right)+\sin\left({x}^2\right)$ & $U\left(-1,1,256\right)$ \\
    Livermore-15 & ${x}^{\frac{1}{5}}$ & $U\left(0,1,256\right)$ \\
    Livermore-16 & ${x}^{\frac{2}{5}}$ & $U\left(0,1,256\right)$ \\
    Livermore-17 & $4\sin\left(x\right) \cos\left(y\right)$ & $U\left(-1,1,256\right)$ \\
    Livermore-18 & $\sin\left({x}^2\right) \cos\left(x\right)-5$ & $U\left(-1,1,256\right)$ \\
    Livermore-19 & ${x}^5+{x}^4+{x}^2+x$ & $U\left(-1,1,256\right)$ \\
    Livermore-20 & $\operatorname{exp}\left({-x}^2\right)$ & $U\left(-1,1,256\right)$ \\
    Livermore-21 & ${x}^8+{x}^7+{x}^6+{x}^5+{x}^4+{x}^3+{x}^2+x$ & $U\left(-1,1,256\right)$ \\
    Livermore-22 & $\operatorname{exp}\left(-0.5{x}^2\right)$ & $U\left(-1,1,256\right)$ \\
    \midrule
    R-1    & $\frac{{\left(x+1\right)}^3}{{x}^2-x+1}$ & $U\left(-1,1,256\right)$ \\
    R-2    & $\frac{{x}^5-3{x}^3+1}{{x}^2+1}$ & $U\left(-1,1,256\right)$ \\
    R-3    & $\frac{{x}^6+{x}^5}{{x}^4+{x}^3+{x}^2+x+1}$ & $U\left(-1,1,256\right)$ \\
    \midrule
    Jin-1 & $2.5 x^4-1.3 x^3 +0.5 y^2 - 1.7y$ & $U\left(-1, 1, 256\right)$ \\
    Jin-2 & $8.0 x^2 + 8.0 y^3 - 15.0$ & $U\left(-1, 1, 256\right)$ \\
    Jin-3 & $0.2 x^3 + 0.5 y^3 - 1.2 y - 0.5 x$ & $U\left(-1.5, 1.5, 256\right)$ \\
    Jin-4 & $1.5 \exp(x) + 5.0 \cos(y)$ & $U\left(-1.5, 1.5, 256\right)$ \\
    Jin-5 & $6.0 \sin(x) \cos(y)$ & $U\left(-1, 1, 256\right)$  \\
    Jin-6 & $1.35 x y + 5.5 \sin((x - 1.0)(y - 1.0))$ & $U\left(-1, 1, 256\right)$ \\
    \midrule
    Koza-2 & $x^5-2x^3+x$ & $U\left(-1, 1, 256\right)$ \\
    Koza-3 & $x^6-2x^4+x^2$ & $U\left(-1, 1, 256\right)$ \\
    \bottomrule 
    \\
\end{longtable}

\section{Extrapolation Experiment}
\label{app:extra}
Figure~\ref{fig:extra} provides a qualitative comparison between the performance of \dysymnet, NeSymReS (pre-training SR method),  MLP (black-box model), and SVR with regards to the ground-truth expression $2sin(1.3x)cos(x)$. The training dataset, represented by the shaded region, comprises 256 data points in the range of [-2, 2], and the evaluation is conducted on the out-of-domain region between [-5, 5]. Although all three models ﬁt the training data very well, our proposed \dysymnet outperforms the NeSymReS baseline in ﬁtting the underlying true function, as shown by the out-of-domain performance. Additionally, the result demonstrates a strong advantage of the symbolic approach: once it has found the correct expression, it can predict the whole sequence, whereas the precision of the black-box model deteriorates as it extrapolates further.

\section{Convergence Analysis}
\label{app:conver}
\begin{figure}[htbp]
    \centering
    \hfill
    \begin{minipage}[t]{0.4\linewidth}
        \includegraphics[width=\linewidth]{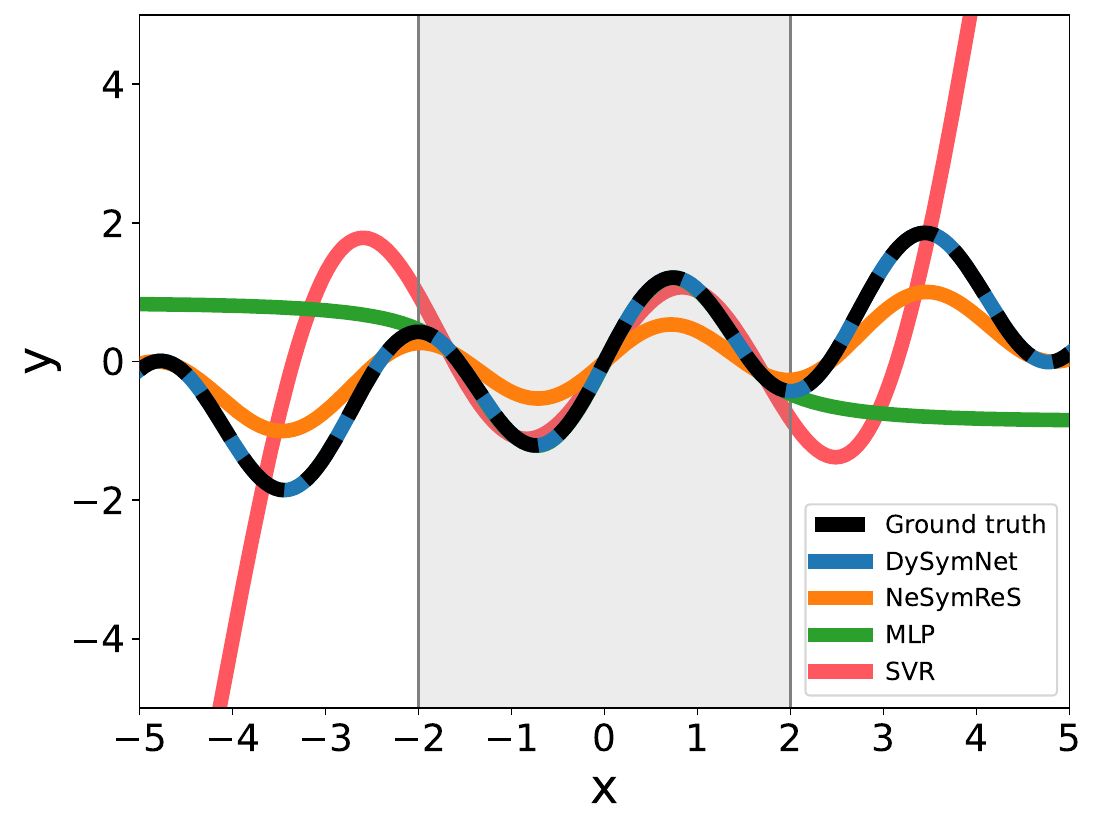}
        \caption{Extrapolation performance comparison with symbolic regression models and black-box models.}
        \label{fig:extra}
    \end{minipage}
    \hfill
    \begin{minipage}[t]{0.4\linewidth}
        \includegraphics[width=\linewidth]{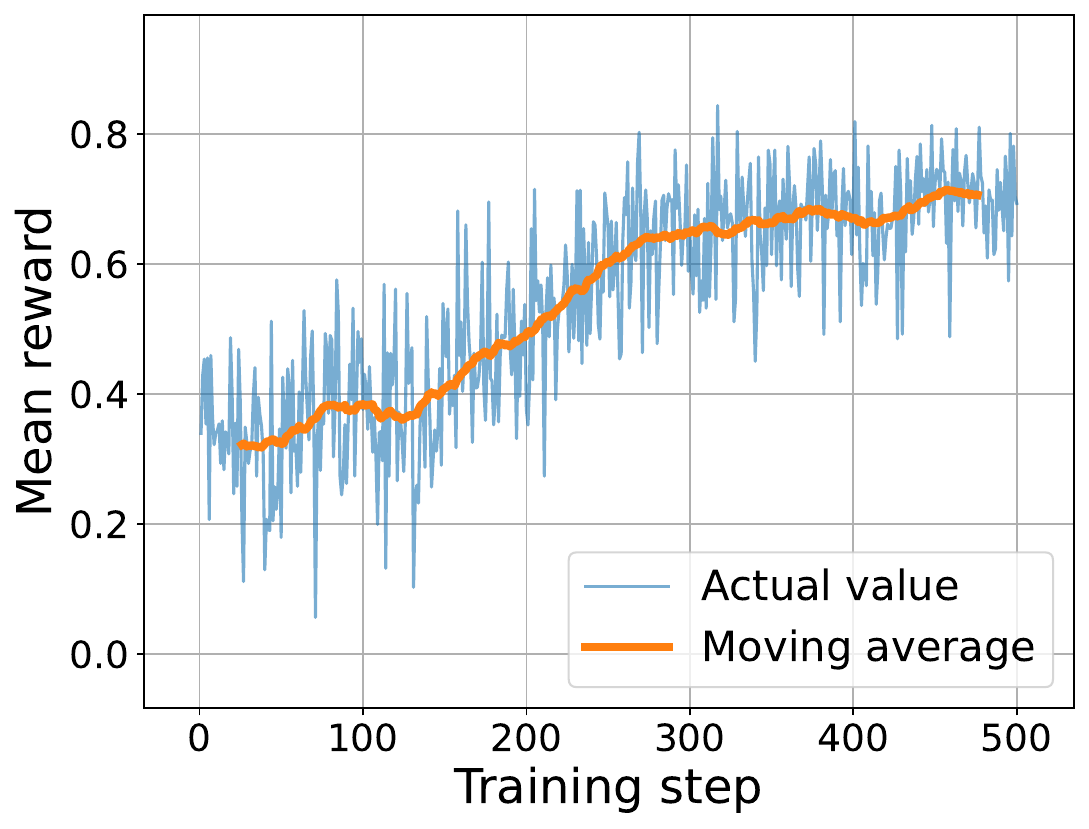}
        \caption{Training curves of RNN for the mean reward of full batch expressions identified from \dysymnet.}
        \label{fig:converagence}
    \end{minipage}  
\end{figure}

In Figure~\ref{fig:converagence}, the average reward computed by the batch expressions, which are identified from \dysymnet, gradually increases with each training step of the policy network RNN, indicating that the direction of the policy gradient is towards maximizing the expected reward value, and the RNN gradually converges to an optimal value.

\section{Free Falling Balls with Air Resistance}
\label{app:balls2}
\begin{figure}[htbp]
  \centering
  \includegraphics[width=\linewidth]{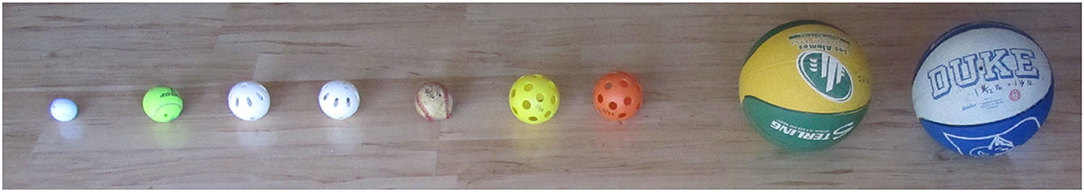}
  \caption{The experimental balls dropped from the bridge~\citep{de2020discovery}. From left to right: golf ball, tennis ball, whiffle ball 1, whiffle ball 2, baseball, yellow whiffle ball, orange whiffle ball, green basketball, and blue basketball. Volleyball is not shown here.}
  \label{fig:balls}
\end{figure}

This section shows more details on exploring the physical laws of free falling balls with air resistance. The experimental ball-drop datasets~\citep{de2020discovery} consist of the records of 11 different types of balls, as shown in Figure~\ref{fig:balls}. These balls were dropped from a bridge and collected at a sampling rate of 30 Hz. Since air resistance affects each ball differently, which in turn leads to different physical laws, we explored the physical laws for each type of ball separately. We split the dataset of each ball into a training set and a test set, where the training set is the recordings from the first 2 seconds and the test set is the recordings from the remaining drop time. 

We explore the physical laws on the training data for each type of ball and then compute the error on the test set. Table~\ref{Table:ball eq} reports the physical laws discovered by \dysymnet and other three baseline physical models. Figure~\ref{fig:balls} illustrates the fitting trajectories of these discovered physical laws.

\begin{figure}[htbp]
  \centering
  \includegraphics[width=\linewidth]{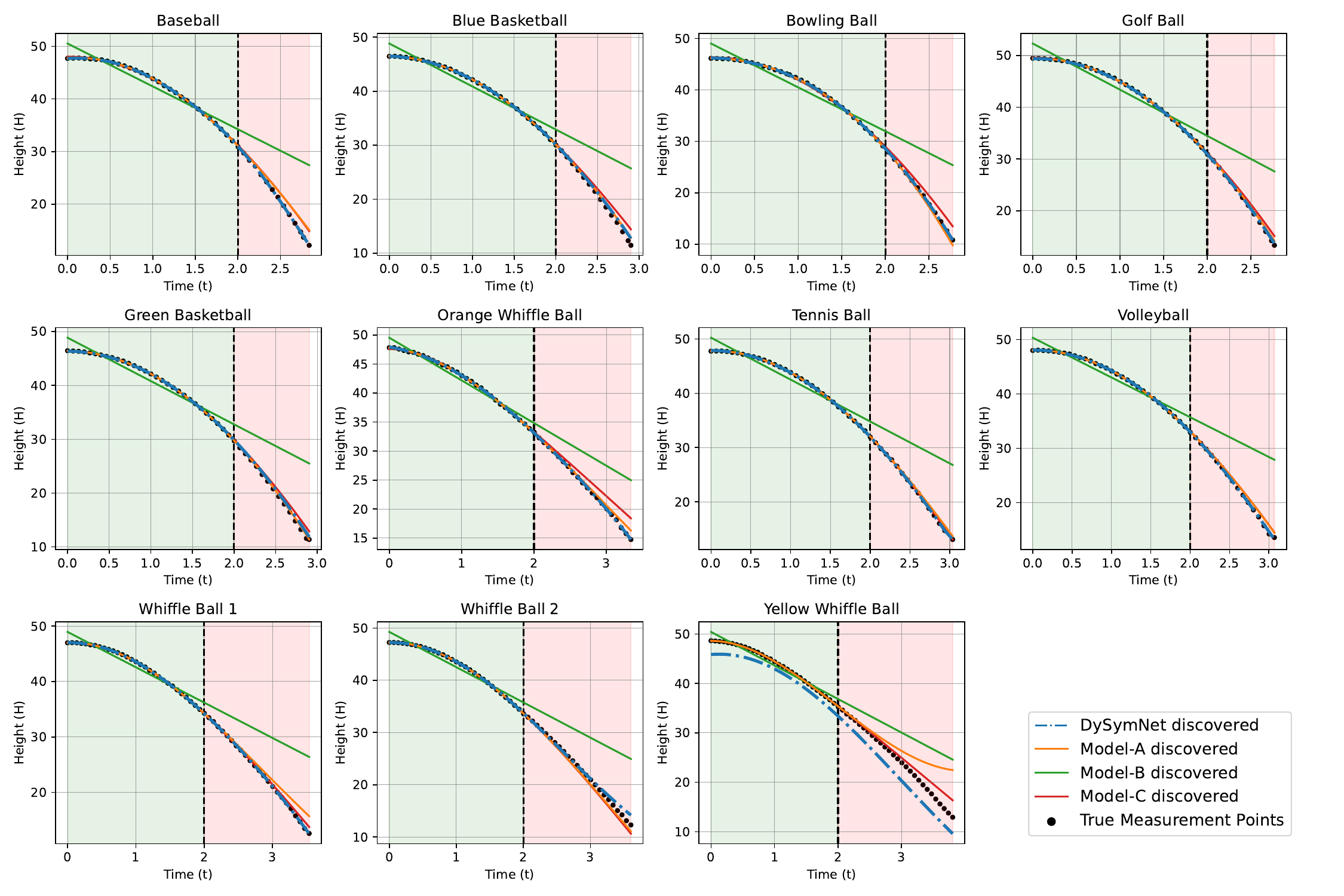}
  \caption{The fitting trajectories of the laws discovered by \dysymnet and three baseline physical models on the dataset of 11 balls. \textcolor{C2}{Green area} represents the extent of the training set (records from the first 2 seconds) and \textcolor{C13}{red area} represents the extent of the test set (records from the remaining drop time).}
  \label{fig:balls res}
\end{figure}

\begin{table}[htbp]
    \centering
    \caption{Discovered physical laws from the motions of free-falling balls by \dysymnet and other three baseline physical models. Note that these formulas are the raw expressions generated by \dysymnet, and further simplification helps to improve the parsimony of the expressions.}
    \vskip 0.1in
    \resizebox{0.8\linewidth}{!}{
    \begin{tabular}{ccl}
    % \small
    % \begin{tabular*}{\linewidth}{@{}LLL@{}}
    \toprule
    \textbf{Ball type} &  \textbf{Method} &  \textbf{Disvovered expression} \\
    \midrule
    Baseball    & \dysymnet & \scriptsize$H(t)=log(cosh(-t*(4.731t - 0.108\sin(4t)) + t + 48.380))$ \\[-0.3ex]
                & Model-A & \scriptsize$H(t)=47.682+1.456t-5.629t^2+0.376t^3$ \\[-0.3ex]
                & Model-B & \scriptsize$H(t)=45.089-8.156t+5.448\exp(0t)$ \\[-0.3ex]
                & Model-C & \scriptsize$H(t)=48.051-183.467\log(\cosh(0.217t))$\\[-0.3ex]
    \midrule
    Blue Basketball      & \dysymnet & \scriptsize$H(t)=(0.588 - 0.172t)*(22.209t + 79.077)$ \\[-0.3ex]
                  & Model-A & \scriptsize$H(t)=46.513-0.493t-3.912t^2+0.03t^3$ \\[-0.3ex]
                & Model-B & \scriptsize$H(t)=43.522-7.963t+5.306\exp(0t)$ \\[-0.3ex]
                & Model-C & \scriptsize$H(t)=46.402-84.791\log(\cosh(0.319t))$\\[-0.3ex]
    \midrule
    Bowling Ball    & \dysymnet & \scriptsize$H(t)=-1.310t(3t+\log(\cosh((\log(\cosh(t))))))$\\[-0.3ex]
    &         & \scriptsize$\quad\quad\quad+\log(\cosh(\log(\cosh(\cos(2.698t)))))+46.111$\\[-0.3ex]
            & Model-A & \scriptsize$H(t)=46.139-0.091t-3.504t^2-0.431t^3$\\[-0.3ex]
                & Model-B & \scriptsize$H(t)=43.336-8.525t+5.676\exp(0t)$\\[-0.3ex]
                & Model-C & \scriptsize$H(t)=46.342-247.571\log(\cosh(0.189t))$\\[-0.3ex]
    \midrule
    Golf Ball   & \dysymnet & \scriptsize$H(t)=(0.592 - 0.182t)*(26.163t + 83.442)$\\[-0.3ex]
                & Model-A & \scriptsize$H(t)=49.413+0.532t-5.061t^2+0.102t^3$\\[-0.3ex]
                & Model-B & \scriptsize$H(t)=46.356-8.918t+5.964\exp(0t)$\\[-0.3ex]
                & Model-C & \scriptsize$H(t)=49.585-178.47\log(\cosh(0.23t))$\\[-0.3ex]
    \midrule
    Green Basketball      & \dysymnet & \scriptsize$H(t)=(0.492 - 0.146t)*(27.582t + 94.230)$ \\[-0.3ex]
     & Model-A & \scriptsize$H(t)=46.438-0.34t-3.882t^2-0.055t^3$ \\[-0.3ex]
                & Model-B & \scriptsize$H(t)=43.512-8.043t+5.346\exp(0t)$ \\[-0.3ex]
                & Model-C & \scriptsize$H(t)=46.391-124.424\log(\cosh(0.263t))$\\[-0.3ex]
    \midrule
    Orange Whiffle Ball    & \dysymnet & \scriptsize$H(t)=-t(t(0.368 - 0.191t^2) + 2t) - t - \exp(t) + 48.929$\\[-0.3ex]
        & Model-A & \scriptsize$H(t)=47.836-1.397t-3.822t^2+0.422t^3$\\[-0.3ex]
        & Model-B & \scriptsize$H(t)=44.389-7.358t+5.152\exp(0t)$\\[-0.3ex]
                & Model-C & \scriptsize$H(t)=47.577-12.711\log(\cosh(0.895t))$\\[-0.3ex]
    \midrule
    Tennis Ball     & \dysymnet & \scriptsize$H(t)=\log(\cosh(3.616t^2 - 33.533)) + \cos(t) + 14.035$\\[-0.3ex]
            & Model-A & \scriptsize$H(t)=47.738+0.658t-4.901t^2+0.325t^3$\\[-0.3ex]
                & Model-B & \scriptsize$H(t)=45.016-7.717t+5.212\exp(0t)$\\[-0.3ex]
                & Model-C & \scriptsize$H(t)=47.874-114.19\log(\cosh(0.269t))$\\[-0.3ex]
    \midrule
    Volleyball  & \dysymnet & \scriptsize$H(t)=t(t(-0.590*t - 4.393) - \cos(t)) + \exp(t) + 47.089$\\[-0.3ex]
                & Model-A & \scriptsize$H(t)=48.046+0.362t-4.352t^2+0.218t^3$\\[-0.3ex]
                & Model-B & \scriptsize$H(t)=45.32-7.317t+5.037\exp(0t)$\\[-0.3ex]
                & Model-C & \scriptsize$H(t)=48.124-107.816\log(\cosh(0.27t))$\\[-0.3ex]
    \midrule
    Whiffle Ball 1     & \dysymnet & \scriptsize$H(t)=\log(\cosh((-1.831t - 6.155)*\log(\cosh(t))$\\[-0.3ex]
    &         & \scriptsize$\quad\quad\quad + \log(\cosh(\log(\cosh(\log(\cosh(t)))))) + 47.719))$\\[-0.3ex]
        & Model-A & \scriptsize$H(t)=46.969+0.574t-4.505t^2+0.522t^3$\\[-0.3ex]
               & Model-B & \scriptsize$H(t)=44.259-6.373t+4.689\exp(0t)$\\[-0.3ex]
                & Model-C & \scriptsize$H(t)=47.062-34.083\log(\cosh(0.462t))$\\[-0.3ex]
    \midrule
    Whiffle Ball 2     & \dysymnet & \scriptsize$H(t)=21.017\sin(0.607t + 7.846) + 26.213$\\[-0.3ex]
         & Model-A & \scriptsize$H(t)=47.215+0.296t-4.379t^2+0.421t^3$\\[-0.3ex]
             & Model-B & \scriptsize$H(t)=44.443-6.744t+4.813\exp(0t)$\\[-0.3ex]
                & Model-C & \scriptsize$H(t)=47.255-38.29\log(\cosh(0.447t))$\\[-0.3ex]
    \midrule
    Yellow Whiffle Ball    & \dysymnet & \scriptsize$H(t)=t + 33.507(0.337\cos(0.365t) + 1)^3 - 34.227$\\[-0.3ex]
         &         & \scriptsize$\quad\quad\quad+48.6092\log(\cosh(t))/(\log(\cosh(t))+3.065)$\\[-0.3ex]
            & Model-A & \scriptsize$H(t)=48.613-0.047t-4.936t^2+0.826t^3$\\[-0.3ex]
                & Model-B & \scriptsize$H(t)=45.443-6.789t+4.973\exp(0t)$\\[-0.3ex]
                & Model-C & \scriptsize$H(t)=48.594-12.49\log(\cosh(0.86t))$\\[-0.3ex]
    \bottomrule
    \end{tabular}
    }
    \label{Table:ball eq}
    % \normalsize
    \vskip -0.1in
\end{table}

\end{document}